\documentclass[a4paper,12pt,times,numbered,print,index]{Classes/PhDThesisPSnPDF}

\input{Preamble/preamble}

\title{On some studies of\\
Fraud Detection Pipeline\\
and related issues from the scope of\\
Ensemble Learning and Graph-based Learning}

\author{Tuan TRAN}

\dept{Quantitative and Computational Finance}

\university{John von Neumann Institute}
\crest{\includegraphics[width=0.2\textwidth]{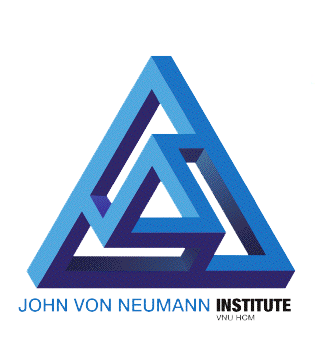}}

\supervisor{Dr. An MAI}

\degreetitle{Master of Science}

\college{JVN Institute}

\subject{LaTeX} \keywords{{LaTeX} {PhD Thesis} {Engineering} {University of
Cambridge}}

\ifdefineAbstract
\fi

\ifdefineChapter
\fi

\begin{document}

\frontmatter

\maketitle


\begin{dedication} 

I would like to dedicate this thesis to my loving parents \dots

\end{dedication}

\begin{declaration}

I hereby declare that except where specific reference is made to the work of 
others, the contents of this dissertation are original and have not been 
submitted in whole or in part for consideration for any other degree or 
qualification in this, or any other university. This dissertation is my own 
work and contains nothing which is the outcome of work done in collaboration 
with others, except as specified in the text and Acknowledgements. This 
dissertation contains fewer than 65,000 words including appendices, 
bibliography, footnotes, tables and equations and has fewer than 150 figures.


\end{declaration}

\begin{acknowledgements}      

I would like to thank Dr. An Mai for his supervision and interesting discussions on several research topics. A great thank you also goes to Loc Tran who helps me complete this thesis.

This work has been partially funded by Vietnam National University - Ho Chi Minh City (VNU-HCM) under grant number C2018-42-02.

\end{acknowledgements}

\begin{abstract}

The UK anti-fraud charity Fraud Advisory Panel (FAP) in their review of 2016 estimates business costs of fraud at £144 billion, and its individual counterpart at £9.7 billion. Banking, insurance, manufacturing, and government are the most common industries affected by fraud activities. Designing an efficient fraud detection system could avoid losing the money; however, building this system is challenging due to many difficult problems, e.g.imbalanced data, computing costs, etc. Over the last three decades, there are various research relates to fraud detection but no agreement on what is the best approach to build the fraud detection system. In this thesis, we aim to answer some questions such as i) how to build a simplified and effective Fraud Detection System that not only easy to implement but also providing reliable results and our proposed Fraud Detection Pipeline is a potential backbone of the system and is easy to be extended or upgraded, ii) when to update models in our system (and keep the accuracy stable) in order to reduce the cost of updating process, iii) how to deal with an extreme imbalance in big data classification problem, e.g. fraud detection, since this is the gap between two difficult problems, iv) further, how to apply graph-based semi-supervised learning to detect fraudulent transactions.

\end{abstract}


\tableofcontents

\listoffigures

\listoftables


\printnomenclature

\mainmatter


\chapter{Introduction}  

\ifpdf
    \graphicspath{{Chapter1/Figs/Raster/}{Chapter1/Figs/PDF/}{Chapter1/Figs/}}
\else
    \graphicspath{{Chapter1/Figs/Vector/}{Chapter1/Figs/}}
\fi

\section{Credit Card Fraud} 

\subsection{Definition and impact of Credit Card Fraud}

A Criminal Code \citep{criminal380} which is a law that codifies most criminal offenses and procedures in Canada, section 380 provides the general definition for fraud in Canada:

\begin{enumerate}
\item Every one who, by deceit, falsehood or other fraudulent means, whether or not it is a false pretense within the meaning of this Act, defrauds the public or any person, whether ascertained or not, of any property, money or valuable security or any service,

\begin{enumerate}
\item is guilty of an indictable offense and liable to a term of imprisonment not exceeding fourteen years, where the subject-matter of the offense is a testamentary instrument or the value of the subject-matter of the offense exceeds five thousand dollars; or
\item is guilty

\begin{enumerate}
\item of an indictable offense and is liable to imprisonment for a term not exceeding two years, or
\item of an offense punishable on summary conviction, where the value of the subject-matter of the offense does not exceed five thousand dollars.
\end{enumerate}

\end{enumerate}

\end{enumerate}

\begin{figure}
\includegraphics[scale=0.8]{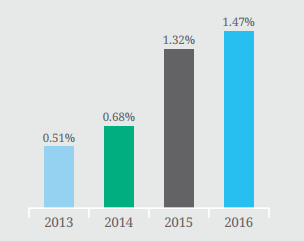}
\centering
\caption{Cost of Fraud as a \% of Revenues Keeps Going Up \citep{lexisnexis2016}}
\label{cost_of_fraud_as_percentage_revenues}
\end{figure}

\begin{figure}
\includegraphics[width=\textwidth]{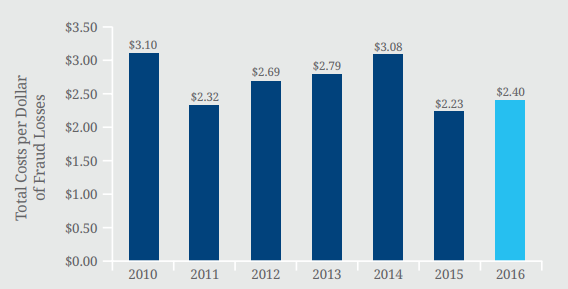}
\centering
\caption{Total Costs per Dollar of Fraud Losses \citep{lexisnexis2016}}
\label{total_cost_per_dollar}
\end{figure}

In recent years, e-commerce has gained a lot in popularity mainly due to the ease of cross-border purchases and online credit card transactions. While e-commerce is already a mature business with many players, security for online payment lags behind. With an extensive use of credit cards, fraud seems like a major issue in the credit card business. It is not easy to see the impact of fraud since companies and banks have a good motivation not to disclose a full report of losses due to frauds. However, as stated in the Lexis Nexis's study \citep{lexisnexis2014}, in 2014 fraudulent card transactions worldwide have reached around \$11 billion a year. A latest study of Lexis Nexis \citep{lexisnexis2016} estimated that a cost of fraud as a percentage of revenues keeps going up, from 0.51\% in 2013 increasing to 1.47\% in 2016 (Figure \ref{cost_of_fraud_as_percentage_revenues}), and a total costs of fraud losses for each dollar is up to 2.4 times (Figure \ref{total_cost_per_dollar}). In the United Kingdom, total losses through credit card fraud have been growing rapidly from £122 million in 1997 to £440.3 million in 2010 according to an estimation of the Association for Payment Clearing Services (APACS) \cite{delamaire2009credit}.

There are many types of fraud, but in this thesis we focus on credit card fraud. The credit card frauds might occur in several ways, for instance:

\begin{itemize}
\item Card-holder-not-present fraud is a payment card transaction made where the card-holder doesn't or can't physically present the card for a merchant's visual examination at the time that an order is given and payment effected, for example a transaction is ordered via mail, telephone or Internet,
\item Stolen card fraud is the most common type of fraud where the stolen card may be used for illegal purchases as much as possible and as quickly as possible until the card holder notifies and blocks the account.
\end{itemize}

\subsection{Credit Card Fraud Detection}

Detecting credit card fraud attracts a lot of attention from industry (bank, insurance, \dots) and from the research community for the last three decades. The fraud detection process is based on the analysis of recorded transactions, which are a set of attributes (for example identifier, transaction timestamp, amount of the transaction, \dots). Traditional methods of data analysis have long been used to detect fraud, it requires large human resource, time-consuming inquiry, wide range of knowledge to analyze a huge number of transactions. However, the human resource is limited and thus automatic fraud detection system is really necessary because it is not easy for a human analyst to detect fraudulent patterns in transaction datasets with a huge number of samples and many attributes. With a rising of Machine Learning field, which is able to learn patterns from data automatically and have been applied in several applications, this is a suitable approach for fraud detection problem. However, a design for the Fraud Detection System (FDS) is particularly challenging for several reasons:

\begin{itemize}
\item The number of credit card frauds is really small - a tiny fraction of all the daily transactions. Many algorithms in Machine Learning have not been designed to deal with this problem,
\item Many fraud transactions have an insignificant difference to normal transactions, which makes these transactions hard to detect by Machine Learning algorithms,
\item Over the time, new attack strategies are created and changes in customer behavior make the distribution of data change over time and it does not meet the basic assumption in Machine Learning (as well as Statistics),
\item \dots.
\end{itemize}

\section{Contributions}

Focusing on the credit card fraud detection problem and some other related issues, in this thesis we present a series of independent contributions through this thesis, including:

\subsection*{Fraud Detection Pipeline}

Despite rich literature on fraud detection, there is no agreement on what is the best way to solve this problem and also there is no public fraud detection system. This means that in the industry, when a company wants to build their own a fraud detection system, they cannot decide which is a good way to do so, especially for those companies in Vietnam. The first main contribution of this thesis is a survey on the fraud detection problem, we will show readers several challenges in building the Fraud Detection System. Based on that, we propose one system that not only good but also simple that companies could implement by themselves easily, which we call Fraud Detection Pipeline. Furthermore, we give readers many possible ways to extend the system.

\subsection*{Update Point Estimation for the case of Concept-Drift in fraud detection problem}

Fraudsters always try to create new fraud strategies and normal customers usually change their behavior, these changes make non-fraud/fraud patterns in our dataset change over time. In this case, the system has to update its models frequently to keep its accuracy, either on a daily or a less frequent period. With the credit card fraud detection problem, a common choice is daily when we receive a daily full dataset and then update the models at night. However, updating models daily is worthless and actually it is only necessary when we do not have to adapt to new fraudulent patterns. In other words, we should update models if and only if it makes an improvement to our system. The second contribution is a mechanism that helps us make a decision on when we should update our models.

\subsection*{K-Segments Under Bagging approach: An experimental Study on Extremely Imbalanced Data Classification}

Imbalanced dataset could be found in many real-world domains, e.g. fraud detection problem, and there are several methods have been proposed to handle the imbalance problem, but there is no guarantee those methods will work with an extreme imbalance case. Big Data is a problem concerning the volume and the complexity of data, whereas the big data in extremely imbalance data is a different problem from those original issues that attracts much attention recently. In this study, as the third contribution, we show that a simplified combination of under-sampling and ensemble learning is very effective in the case mentioned.

\subsection*{Solve fraud detection problem by using graph based learning methods}

To detect the credit cards' fraud transactions, data scientists normally employ the un-supervised learning techniques and supervised learning technique. Among several algorithms that have been proposed for this problem, a graph-based semi-supervised learning have not been applied to the credit cards' fraudulent transactions detection before. We propose to apply an un-normalized graph p-Laplacian based semi-supervised learning technique combined with an undersampling technique to find a relationship of those frauds in the data.

\section{Outline}

This thesis is organized as follows:

\begin{itemize}
\item Chapter 2 prepares the necessary knowledge which would be helpful for the next chapters,
\item Chapter 3 reviews several challenges of the fraud detection problem and their possible solutions, based on that we select the most suitable approaches and combine them into one best strategy as in our proposed Fraud Detection Pipeline,
\item Chapter 4 uses the Fraud Detection Pipeline and presents a mechanism to detects which model need to update to keep the accuracy of our system and also reduces the update cost,
\item Chapter 5 explains more details of our selected techniques in the pipeline how they are useful to us in the case of the extremely imbalanced big dataset,
\item And finally, chapter 7 summarizes our results and proposes future research directions.
\end{itemize}


\chapter{Preliminaries}

\ifpdf
    \graphicspath{{Chapter2/Figs/Raster/}{Chapter2/Figs/PDF/}{Chapter2/Figs/}}
\else
    \graphicspath{{Chapter2/Figs/Vector/}{Chapter2/Figs/}}
\fi

\section{Machine Learning}
\label{preliminaries:machine_learning}

Machine learning is a field in computer science that we try teaching the computers to do tasks without explicitly programmed \citep{samuel2000some, koza1996automated}. Machine learning is strongly associated with computational statistics, which explores data and builds algorithms that could find patterns in data and make predictions. More formal definition of the algorithms used in the machine learning is provided by Mitchell ``A computer program is said to learn from experience E with respect to some class of tasks T and performance measure P if its performance at tasks in T, as measured by P, improves with experience E`` \citep{mitchell1997machine}.

Typically, tasks in machine learning can be grouped into three wide categories \citep{russell1995modern}:

\begin{itemize}
\item Supervised learning: the given data contains input values and their desired output values, the aim of algorithms in supervised learning is to find a general rule that maps inputs to outputs. In this thesis we are focus on the supervised learning approach for our fraud detection problem,
\item Unsupervised learning: there are no desired outputs in the dataset, it means that we give the learning algorithms freely to explore a hidden structure of the data,
\item Reinforcement learning: this kind of learning is slightly different with the above, reinforcement learning builds a dynamic environment then let a computer interacts with it to perform a certain goal, after each action a feedback is provided to the computer as a reward or a punishment.
\end{itemize}

Supervised learning is a most common task in the world and can be seen as a function from labeled training data \citep{mohri2012foundations}. Given a set of $N$ data points of a form ${(x_1, y_1), \dots, (x_N, y_N)}$ such that $x_i$ is a attribute vector of the i-th data point and $y_i$ is its desired output, a supervised learning algorithm is a function $g: X \rightarrow Y$, where $X$ is an input space (matrix of attribute vectors) and $Y$ is a output space (matrix of desired outputs). Typically the function $g$ is in a function space $G$, or we could represent $g$ as a scoring function $f: X \times Y \rightarrow \mathbb{R}$ such that function $f$ will return the output value $y$ which has the highest score $g(x) = argmax_y f(x, y)$. For example, $g$ could be a conditional probability model $g(x) = P(y \mid x)$, e.g. logistic regression \citep{walker1967estimation, cox1958regression}, or $f$ could takes a form of a join t probability model $f(x, y) = P(x, y)$, e.g. naive Bayes \citep{russell1995modern}.

In order to measure a performance of learning function, we could define a loss function $L: Y \times Y \rightarrow \mathbb{R}$ and a risk $R(g)$ of learning function $g$ is defined as an expected loss of function $g$ which could be estimated by:

\begin{equation}
\widehat{R} (g) = \dfrac{1}{N} \sum_i L(y_i, g(x_i))
\end{equation}

\subsection*{Decision tree learning}

\begin{figure}
\includegraphics[scale=0.4]{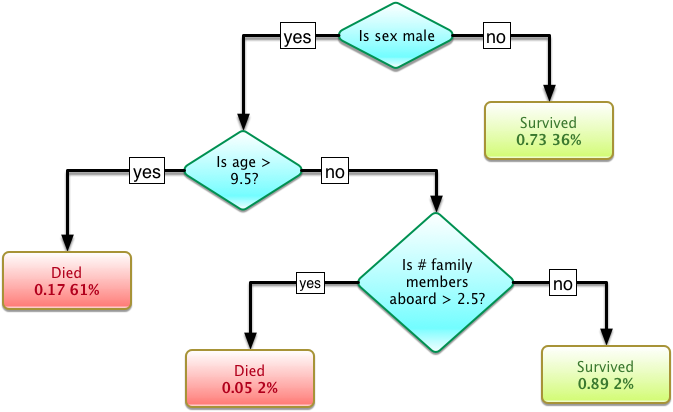}
\centering
\caption{A tree showing survival of passengers on the Titanic}
\label{img:titanic_survival_decision_tree}
\end{figure}

A decision tree learning is one common approach in data mining \citep{rokach2014data}, it aims to create a model that predicts values like making a decision from a tree-based analysis. An example is shown in the figure \ref{img:titanic_survival_decision_tree} \citep{decision_tree_learning}.

A tree learner could classify the data points by splitting the input dataset into subsets based on a given criterion. The splitting process at each node is repeated in a recursive procedure until all data points are classified. There are two kinds of decision trees:

\begin{itemize}
\item Regression tree: create a tree to predict the desired output as a real number,
\item Classification tree: an analysis aims to predict a categorical output.
\end{itemize}

A term Classification and Regression Tree (CART) analysis is used to refer to both the above analysis \citep{breiman1984classification}. There are many decision tree algorithms, including:

\begin{itemize}
\item ID3 (Iterative Dichotomiser 3) \citep{quinlan1986induction},
\item C4.5: an extension of ID3 algorithm \citep{quinlan2014c4}.
\end{itemize}

Algorithms for constructing decision trees use different metrics for measuring which is the best feature to split the dataset. These measures compute the homogeneity of the output values within the subsets, some common measures in decision tree learning is Entropy, Information Gain, and Gini.

\subsection*{Ensemble learning}

In machine learning, ensemble learning is a method combines multiple learning algorithms to obtain better accuracy than from any individual algorithms \citep{opitz1999popular, rokach2010ensemble}. The term ensemble is usually reserved for methods that generate multiple hypotheses using the same base learner. There are three common types of ensemble learning, such as:

\subparagraph{Bootstrap aggregating (bagging)}

\begin{figure}
\includegraphics[scale=0.4]{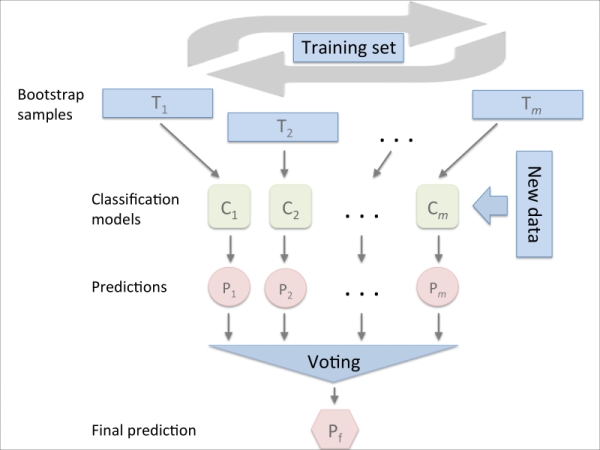}
\centering
\caption{Bagging ensemble \citep{raschka2015python}}
\label{img:bagging_ensemble}
\end{figure}

An ensemble technique is bootstrap aggregating which is usually called bagging (see figure \ref{img:bagging_ensemble}), it combines multiple learnings via their equal weight votes. In order to support the model variance, bagging trains each model from a randomly drawn subset. And a famous example of bagging is the Random Forest algorithm which uses random decision trees as base learners.

\subparagraph{Boosting}

\begin{figure}
\includegraphics[scale=0.4]{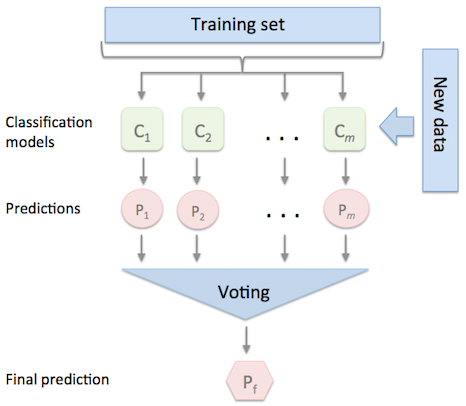}
\centering
\caption{Boosting ensemble \citep{raschka2015python}}
\label{img:boosting_ensemble}
\end{figure}

Similar with the bagging ensemble, boosting is an incremental algorithm that will train each new model to underline the data points that previous models mis-classified (see figure \ref{img:boosting_ensemble}). Boosting could have a better accuracy than bagging in some cases, however it also tends to overfit to the training data. A most common boosting algorithm is Adaboost \citep{freund1999short}.

\subparagraph{Stacking}

\begin{figure}
\includegraphics[scale=0.4]{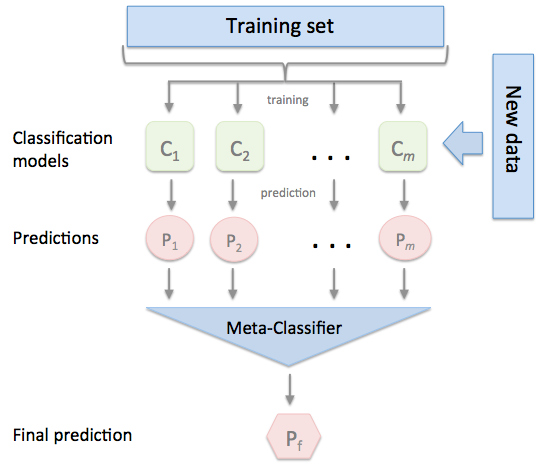}
\centering
\caption{Stacking ensemble \citep{raschka2015python}}
\label{img:stacking_ensemble}
\end{figure}

Slightly different with these above techniques, stacking is a supervised learning algorithm that uses the predictions of several other learning algorithms as an input dataset to makes the final prediction (see figure \ref{img:stacking_ensemble}). All of the other algorithms are trained using the given dataset independently, then a combiner algorithm is trained on that which can be any algorithm. Therefore, stacking can theoretically represent any of the ensemble techniques, although in practice we often use the Logistic Regression or Linear Regression as the combiner.

\subsection*{Random Forest}

Random Forest \citep{breiman2001random} or random decision forests \citep{ho1995random, ho1998random} are an ensemble learning method for classification or regression that perform by constructing many trees, e.g. decision trees, then combine these results to makes final output values as a mode of the classes for classification tasks or a mean prediction for regression tasks. The Random Forest model based on two elements:

\begin{enumerate}
\item Base learner: each tree in the forest is the base learner, typically it is a weak learner with high variance,
\item Ensemble learning: combine the results of several weak learners to make a final prediction.
\end{enumerate}

The Decision Tree above that are grown very deep tend to overfit their training sets, i.e. have a low bias, but very high variance. We use the tree-based method as the base learner by its strength, and to handle its weakness we combine all of the predictions via ensemble learning. This method is called Random Forest, it can formulate the algorithm as the following statement: given a dataset $X = x_1, \dots, x_n$ with $n$ data points and respectively desired output values $Y = y_1, \dots, y_n$, bagging repeatedly $B$ times selects a random sample from the dataset $X$ then training trees on these samples.

\begin{algorithm}
   \caption{Tree bagging}
    \begin{algorithmic}
      \Function{bagging}{$X, Y, B$}

        \For{$b = 1$ to ${B}$}
            \State $(X_b, Y_b)$ is samples from $(X, Y)$ which have $n$ data points
            \State Train a classification or regression tree $f_b$ on $X_b, Y_b$
        \EndFor
        
       \EndFunction

\end{algorithmic}
\end{algorithm}

The authors of Random Forest have proposed using an average function to makes the final prediction in regression tasks or takes the majority vote in classification tasks \citep{breiman2001random}. However, there is no standard function as a combiner, thus we are free to choose the combination function. This bootstrapping procedure could decrease the variance of the model without increasing the bias. In order to estimate the uncertainty of the prediction, it could be computed as a standard deviation of the predictions from all the individual trees:

\begin{equation}
\sigma = \sqrt{ \dfrac{ \sum_{b=1}^B ( f_b(x) - \hat{f} ) }{ B - 1 } }
\end{equation}

Random Forest uses the above bagging algorithm with only one difference, it uses a modified tree that selects a random subset of the features to split a current subset data. Typically, the dataset has $p$ features then for a classification problem $\sqrt{p}$ features will be used in each split and for regression problems the inventor recommends that we should use $p/3$ as the number of features \citep{friedman2001elements}.

\section{Measurement}
\label{measurement_premilinaries}

The fraud detection problem is the classification task and in order to measure how well of the classifiers, there are many metrics could be used to evaluate the algorithms. In this section, we will review some common metrics which are typically used in classification tasks.

\subsection*{Accuracy}

Accuracy is a measure of statistical variablity which represent a percentage of correct predictions on the total number of cases examined. In the fields of science and engineering, the accuracy of a measurement system is the degree of closeness of measurements of a quantity to that quantity's true value \citep{bipm2008international}. Consider 2 sets with $n$ data points: the true labels from the given dataset $Y = {y_1, \dots, y_n}$ and our predicted values $\hat{Y} = {\hat{y_1}, \dots, \hat{y_n}}$, the accuracy is:

\begin{equation}
accuracy = \dfrac{ 1 }{ n } \sum_{i = 1}^n \mbox{1\hspace{-4.25pt}\fontsize{12}{14.4}\selectfont\textrm{1}}_{ y_i = \hat{y_i} }
\end{equation}

\subsection*{Precision and recall}

\begin{figure}
\includegraphics[scale=0.35]{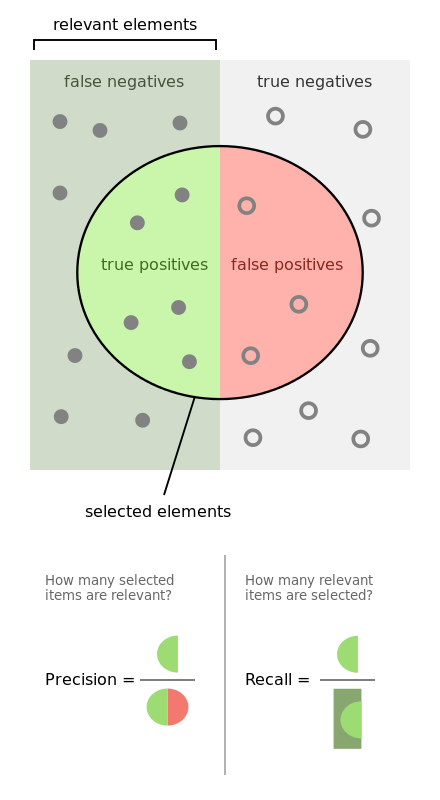}
\centering
\caption{Precision and recall \citep{wiki2018precision_recall}}
\label{img:precision_and_recall}
\end{figure}

In pattern recognition, information retrieval and binary classification, precision and recall are two common metrics. Precision (also called positive predictive value) is the fraction of relevant data points among the given data points, and recall (also known as sensitivity) is the fraction of relevant data points that have been retrieved over the total amount of relevant data points (see figure \ref{img:precision_and_recall}).

In the information retrieval contexts, precision and recall were defined by Perry, Kent \& Berry (1955) \citep{perry1955machine} as a set of retrieved elements and a set of relevant elements. Precision is the fraction of retrieved documents that are relevant to the query:

\begin{equation}
\text{precision} = \dfrac{ \mid {\text{relevant elements}} \cap {\text{retrieved elements}} \mid }{ \mid {\text{retrieved elements}} \mid }
\end{equation}

And the recall is the fraction of the relevant elements that are successfully retrieved:

\begin{equation}
\text{precision} = \dfrac{ \mid {\text{relevant elements}} \cap {\text{retrieved elements}} \mid }{ \mid {\text{relevant elements}} \mid }
\end{equation}

In the classification tasks, considers four terms true positives (TP), true negatives (TN), false positives (FP), and false negatives (FN). Precision and recall are defined as \citep{olson2008advanced}:

\begin{equation}
\text{precision} = \dfrac{ TP }{ TP + FP }
\end{equation}

\begin{equation}
\text{recall} = \dfrac{ TP }{ TP + FN }
\end{equation}

\subsection*{Receiver operating characteristic (ROC)}

\begin{figure}
\includegraphics[scale=2.5]{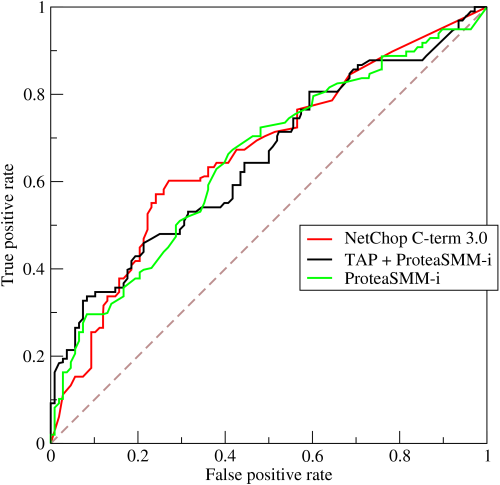}
\centering
\caption{ROC curve of three predictors of peptide cleaving in the proteasome \citep{wiki2018roc}}
\label{img:roc_curve}
\end{figure}

A receiver operating characteristic curve is a graphical plot which is created by plotting the true positive rate (TPR) against the false positive rate (FPR) at various threshold settings. ROC analysis provides tools to select a possible optimal threshold for models, an example ROC curve plot of three predictors of peptide cleaving in the proteasome is shown in figure \ref{img:roc_curve}.

\subsection*{Area under the ROC curve (AUC)}

We cannot compare classifiers base on the ROC because the ROC curve is just a graphical plot. In order to represent ROC performance as a single scala, a common method is to calculate the area under the ROC curve (AUC for short). Because the AUC is a size of the area of the unit square, therefore the AUC value will always be between 0 and 1. However, random guessing produces an area of 0.5, no realistic classifier should have an AUC less than 0.5 then it makes the baseline performance for all classifiers is 0.5. In summary, the AUC metric is a measure of how much the ROC curve is close to the point of perfect classification.

The AUC metric usually used in machine learning community for model comparison \citep{hanley1983method}. However, in practice researchers recently noticed that AUC was quite noisy as a classification measure \citep{hanczar2010small} and some other studies have other significant problems in model comparison \citep{lobo2008auc, hand2009measuring}.

\subsection*{F-score}

$F_1$ score (also F-score or F-measure) is a measure that considers both precision and recall by taking a harmonic average of these metrics.

\begin{equation}
F_1 = 2 \cdot \dfrac{ \text{precision} \cdot \text{recall} }{ \text{precision} + \text{recall} }
\end{equation}

A general formula of F-score for any positive real $\beta$ is:

\begin{equation}
F_\beta = (1 + \beta^2) \cdot \dfrac{ \text{precision} \cdot \text{recall} }{ (\beta^2 \cdot \text{precision}) + \text{recall} }
\end{equation}

Van Rijsbergen et al. \citep{van1979information} interpret $F_\beta$ that "measures the effectiveness of retrieval with respect to a user who attaches $\beta$ times as much importance to recall as precision". In practice, the F-score metric is often used in machine learning when the dataset is imbalanced because it isn't affected by the imbalance problem.
\chapter{Fraud Detection Pipeline}
\label{fraud_detection_pipeline}

\ifpdf
    \graphicspath{{Chapter3/Figs/Raster/}{Chapter3/Figs/PDF/}{Chapter3/Figs/}}
\else
    \graphicspath{{Chapter3/Figs/Vector/}{Chapter3/Figs/}}
\fi

\vspace*{\fill}
\thispagestyle{empty}
\epigraph{Standing on the shoulders of giants }{ --- Bernard of Chartres}
\thispagestyle{empty}

There are various research and studies on fraud detection problem and related issues,  including what algorithms should be used \citep{leonard1993detecting, bolton2001unsupervised, mahmoudi2015detecting}, how to prepare features \citep{bahnsen2016feature, whitrow2009transaction, fast2007relational}, \dots. In this section, we will review challenges in building the Fraud Detection System and its relevant questions. Based on that, we propose a Fraud Detection Pipeline which is not only simple, reliable but also easy to implement. Furthermore, we provide the readers many potential improvements for the pipeline.

\section{Imbalanced dataset}
\label{imbalanced_dataset}

One of the main problems of fraud detection is dataset imbalanced, i.e. the number of fraudulent transactions are a small fraction of the dataset \citep{juszczak2008off}. The imbalanced problem is of significant concern in the data mining and machine learning community because imbalanced datasets are common in many real-world domains. For instance, in the detection of fraudulent cases in telephone calls \citep{fawcett1997adaptive} and credit card transactions \citep{chan1999distributed}, the number of legitimate transactions heavily outnumbers the number of fraudulent transactions. Learning from imbalanced data sets is an important issue in supervised learning, with the credit card fraud detection problem, the imbalance is a big problem for the learning process. Let consider an example which consists only 1\% fraudulent transactions (minority class) and the remaining belongs to legitimate category (majority class), a lazy classification that predicts all of the transactions are legitimate transactions will has 99\% accuracy. This is a very high accuracy but we can not use this because it cannot detect any fraudulent transactions at all.

There are several methods that have been proposed to deal with the imbalanced problem and these could be separated into two levels: methods that operate at the data level and at the algorithmic level \citep{chawla2004special}. Algorithms at the algorithmic level are designed to deal with the minority class detection and at the data level, the rebalanced strategies are used as a pre-processing step to rebalance the dataset before any algorithm is applied.

\subsection*{Algorithmic level methods}
\label{algorithmic_level_methods}

At algorithmic level, these algorithms are modified or extended of existing classification algorithms for imbalanced tasks. Based on their styles we can separate these algorithms into two styles: imbalanced learning and cost-sensitive learning. In the first learning, the algorithms try to improve the accuracy of the minority class prediction, while the second learning tries to minimize the cost of wrong predictions.

\subsubsection*{Imbalance learning}
\label{imbalance_learning}

Decision tree, e.g. C4.5 \cite{quinlan2014c4}, use Information Gain as splitting criteria to maximize the number of predicted instances in each node, it makes the tree bias towards the majority class. Cieslak and Chawla \citep{cieslak2008learning} suggest splitting with Hellinger Distance (HD) which they show that HD is skew-insensitive and their proposed Hellinger Distance Decision Tree is of better performance compared to the standard C4.5. Other studies have also reported the negative effect of skewed class distributions not only in decision tree \citep{he2009learning, japkowicz2002class}, but also in Neural Network \citep{japkowicz2002class, visa2005issues}, k-Nearest Neighbor (kNN) \citep{kubat1997addressing, mani2003knn, batista2004study} and SVM \citep{yan2003predicting, wu2003class}.

In Machine Learning, with a great success of ensemble learning on several applications, many ensemble strategies have been proposed for imbalanced learning. Bagging \citep{breiman1996bagging} and Boosting \citep{freund1996experiments} are the most popular strategies which combine the imbalanced strategy with a classifier to aggregate classifiers \citep{liu2009exploratory, wang2009diversity, vilarino2005experiments, kang2006eus, liu2006boosting, wang2010boosting, chawla2003smoteboost, joshi2001evaluating, mease2007boosted}.

\subsubsection*{Cost-sensitive learning}
\label{cost_sensitive_learning}

In classification applications dealing with imbalanced datasets, the correct prediction of minority class is more important than the correct prediction of majority class, which causes several classifiers to fail when predicting minority class because they assume the cost of these classes are the same. In the credit card fraud problem, the cost of our true prediction is zero, but if we cannot detect a fraud transaction then its amount of money is our lost and if we predict a non-fraud is a fraud then the cost is an investigation fee need to correct this transaction (see table \ref{tab:simple_cost_matrix} for a simple cost matrix).

\begin{table}[h!]
  \centering
  \caption{A simple cost matrix for one transaction.}
  \label{tab:simple_cost_matrix}
  \begin{tabular}{|c|c|c|}
      \hline
    &Non-fraud&Fraud \\
    \hline
    Predict non-fraud&0&its amount of money \\ \hline
    Predict fraud&investigation fee&0 \\ \hline
  \end{tabular}
\end{table}

Classifiers in cost-sensitive learning use different costs for prediction of each class, these cost-based classifiers could handle wrong-prediction costs without sampling or modifying the dataset. For example in tree-based classifiers, cost-based splitting criteria are used to minimize costs \citep{ling2004decision}; or used in tree pruning\citep{bradford1998pruning}. An extension of cost-sensitive learning, Domingos proposed Metacost \citep{domingos1999metacost} framework that transforms non-cost-sensitive algorithm into a cost-sensitive algorithm. However, the cost for minority class is usually not available or difficult to compute, which makes the cost-sensitive algorithms are not popular \citep{maloof1997learning}.

\subsection*{Data level methods}
\label{data_level_methods}

Methods at data level are techniques that modifying an imbalanced dataset before any classifiers could be applied. In this section, we will introduce some popular techniques usually used to re-balance the distribution of the classes.

\subsubsection*{Sampling}
\label{sampling}

\begin{figure}[!ht]
\centering
\caption{Three common sampling methods \citep{dal2013racing}}
\label{sampling_methods}
\includegraphics[]{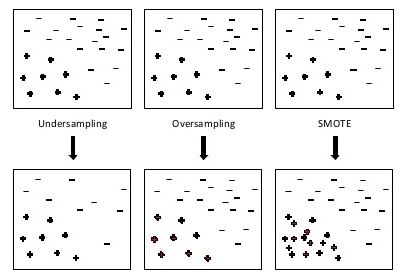} 
\end{figure}

Several studies \citep{weiss2001effect, laurikkala2001improving, estabrooks2004multiple} have shown that balanced training set will give a better performance when using normal algorithms, and sampling methods are the most common techniques in data science to create a balanced training set. There are three popular sampling techniques, including undersampling, oversampling and SMOTE (see figure \ref{sampling_methods}); and some extension sampling techniques are based on these three techniques, e.g. Borderline-SMOTE \citep{han2005borderline}, ADASYN \citep{he2008adasyn}, \dots.

Undersampling \citep{drummond2003c4} is a method to decrease the size of the majority class by removing instances at random. The idea is that many instances of the majority class are redundant and the removal of these instances, thereby, makes its distribution change not too much. However, the risk of dropping redundant instances still exists since this process is unsupervised and we cannot control what instances will be dropped. Furthermore, a perfectly balanced dataset, which means the size of the minority class equals the size of the majority class, is not a good choice for undersampling \citep{dal2015calibrating}. In practice, this technique is often used because it is simple and speeds up our process.

Oversampling \citep{drummond2003c4} is an opposite method of undersampling that tries to increase the size of minority class at random. While duplicating the minority class, oversampling increases a risk of overfitting by biasing the classifier toward the minority class \citep{drummond2003c4}. Furthermore, this method does not add any new information for minority instances and it also slows down our learning phase.

SMOTE \citep{chawla2002smote} is a small extended version that combines both sampling techniques above, abnormal oversampling and normal undersampling. While oversamples the minority class by creating similar instances in a neighborhood area and also do undersampling the majority instances randomly, it could create clusters around each minority instance and helps classifiers build larger decision regions. SMOTE has shown to increase the accuracy of classifiers \citep{chawla2002smote}, a change of learning time depends on sampling ratio but it still has some drawbacks, e.g. new minority instances are created without considering its neighborhood so it increases the overlapping area between the classes. \citep{wang2004imbalanced}.

\subsubsection*{Cost-based}
\label{cost_based}

Cost-based methods is a kind of sampling technique that considers the misclassification cost which is assigned to each instance a different value. The above sampling techniques are random, with cost-based sampling methods, the weight of minority class is usually higher than the weight of majority class and then other sampling techniques could be used to rebalance the datasets.

Costing \citep{zadrozny2003cost} is a cost-based undersampling that draws instances with an acceptable probability greater than a pre-defined threshold. Klement et al. \citep{klement2009cost} proposed a more complicated and effective approach that combines cost-based random under-sampling and ensemble learning. The cost-based methods are promising when dealing with imbalanced datasets, however, the cost of misclassification for each class may be not easy to see in the real life as we have mentioned in the section above.

\subsubsection*{Distance-based}
\label{distance_based}

Distance-based methods are slightly different than cost-based methods, i.e. instead of considering the cost of misclassification. These methods consider a distance among instances in the imbalanced dataset to undersample or to remove a noise and borderline instances of each class. They need to compute the distance between instances so these methods are more costly than the above methods.

Tomek \citep{tomek1976two} proposed a method that removes instances from the majority class that is close to the minority region in order to have a better separation between two classes. His method is useful in noisy datasets or datasets with the overlapping problem, i.e. removing those instances can make the classifier toward misclassification \citep{suman2005methods}. There are several distance-based sampling studies, e.g. Condensed Nearest Neighbor \citep{hart1968condensed}, One Sided Selection \citep{kubat1997addressing}, Edited Nearest Neighbor \citep{wilson1972asymptotic} or Neighborhood Cleaning Rule \citep{laurikkala2001improving}.

\section{Algorithms}
\label{algorithms}

In the fraud detection problem, almost Machine Learning algorithms, including supervised learning \citep{chan1999distributed, kubat1997addressing, bhattacharyya2011data}, unsupervised learning \citep{tasoulis2006unsupervised, bolton2001unsupervised}, ..., have been proposed to detect fraudulent transactions. In this section, we will see which algorithms are used to solve the fraud detection problem.

\subsection*{Expert Systems}

In the very early 90s, the decision which decides a transaction is a fraud or not follows from rules and the rules are generated from the knowledge of human expert. Expert's rules are just simple IF-THEN rules and extremely manually since all of the rules based on the expert's knowledge and ability. By applying expert system, suspicious activity or transaction can be detected from deviations from normal spending patterns \cite{leonard1995development}. Kevin at el. proposed expert system model to detect fraud for alert financial institutions \cite{leonard1993detecting}. Nik presented a FUZZY system to detect credit card frauds in various payment channels, their model fuzzy expert system gives the abnormal degree which determines how the new transaction is fraudulent in comparison with user behavioral \cite{haratinik2012fuzzgy}.

\subsection*{Association Rules}

Association rules are an extended approach of Expert Systems in which we add a certainty factor for expert's rules. Usual measures for the certainty are proposed by Agrawal et al \cite{agrawal1993mining} for establishing an association rule’s fitness and the interest
are the confidence $Conf(A \Rightarrow C)$, the conditional probability $p(C \mid A)$, the support $Supp(A \Rightarrow C)$, and the joint probability $p(A \cup C)$, \dots.

There are few studies use association rules and its extension, e.g. fuzzy association rules \cite{sanchez2009association}. However, the expert's rules are still generated manually and it is the biggest disadvantage of those methods. In the current computing age, a complexity of the data is increasing fastly and we need to create rules or make fraud prediction automatically.

\subsection*{Neural Networks}

The neural networks are non-linear statistical data modeling tools that are inspired by the functionality of the human brain using a set of interconnected nodes \citep{yeh2009comparisons, ghosh1994credit}. Because of its strengths, neural networks are widely applied in various applications such as automobile insurance and corporate fraud. The literature describes that neural networks can be used as a financial fraud detection tool. The neural network fraud classification model employing endogenous financial data created from the learned behavior pattern can be applied to a test sample \citep{green1997assessing}. The neural networks can be used to predict the occurrence of corporate fraud at management level \citep{cerullo1999using}.

Researchers have explored the effectiveness of neural networks, decision trees and Bayesian belief networks in detecting fraudulent financial statements (FFS) and to identify factors associated with FFS \citep{kirkos2007data}. The study in \citep{fanning1998neural} revealed that input vector consisted of financial ratios and qualitative variables, was more effective when fraud detection model was developed using the neural network. The model was also compared with standard statistical methods like linear and quadratic discriminant analysis, as well as logistic regression methods \citep{fanning1998neural}.

The Bayesian belief network (BBN) is a graphical model that presents a probabilistic dependency using a directed acyclic graph (DAG), in which nodes represent random variables and missing edges encode conditional independencies between the variables \citep{kirkos2007data}. The Bayesian belief network is used in developing models for the credit card, automobile insurance, and corporate fraud detection. Bayesian belief network outperformed neural network and decision tree methods and achieved outstanding classification accuracy \citep{kirkos2007data}.

\subsection*{Logistic Regression (LR)}

Logistic Regression \cite{walker1967estimation, cox1958regression} is an important machine learning algorithm. The goal of LR is to model the probability of a target belong to each class. Logistic Regression is a simple and efficient approach that is a widely used technique in such problems \cite{hosmer2013applied}. There are many studies tried to use logit models to estimate fraudulent probability in the case of insurance frauds \cite{jin2005binary, artis2002detection} and other related areas like food stamp programs, and so forth \cite{bollinger1997modeling, hausman1998misclassification, poterba1995unemployment}.

\subsection*{Decision Tree}

A Decision Tree is a decision support tool that uses a tree-like graph to find a prediction rule from the labeled dataset. In our fraud detection case, this means that it could generate expert-rules IF-THEN automatically. The Decision Tree has many advantages, e.g. simple to understand and interpret, important insights can be generated automatically like experts, \dots and it has found several applications in fraud detection \citep{csahin2011detecting, sahin2013cost, bahnsen2015example, lee2006mining}.

\subsection*{Random Forest}

Random Forest \citep{breiman2001random} is an ensemble of Decision Trees, where each tree is trained on a different bootstrap sample of the original training set and uses a random subset of all the features available. The forest of Decision Trees that are very different from each other, this diversity is a key factor for variance reduction in order to overcome the disadvantage of Decision Tree \citep{krogh1995neural}.

Several studies have shown that it achieves the best results among different classifiers \citep{dal2014learned, dal2015credit, van2015apate, whitrow2009transaction, bhattacharyya2011data, bahnsen2013cost} when using Random Forest. Pozzolo et al. \citep{dal2013racing} have shown that the Random Forest combined with undersampling or SMOTE are often the best choice. In our case, we refer undersampling over SMOTE since our data is usually huge. We propose using Random Forest with undersampling for many reasons, e.g. powerful, fast, simple to interpret, \dots.

\section{Sample Selection Bias}
\label{sample_selection_bias}

A standard assumption, not only in Data Science but also in Statistics, is stationary, i.e. the training set and testing set have the same distribution, since these sets come from one generating source. In a non-stationary environment, there is a distributional mismatch between the training set and testing set and classifiers are fitted using the training set could not use the testing set to check its accuracy.

This situation could occur when we modify the original dataset, e.g. under-sampling, and a newly created training set doesn't represent well for the whole dataset. We called this a Sample Selection Bias (SSB), which also happens in the collecting data step when we do not or could not collect data that present the population distribution. For example, when predicting a new customer that could pay a loan (and its interest) at the end of this month and we give a try with a current data of a bank, however the bank only records old customers that repay and they directly refuse a new bad customer without recorded information.

Let $s$ is a random variable which takes either 1 (sampled) or 0 (not sampled). Using Bayes formula we could re-write the $P(x,y)$ as:

\begin{equation}
P(x, y) = \dfrac{ P(x, y|s = 1)P(s = 1) }{ P(s = 1|x, y) } = \dfrac{ P(s = 1) }{ P(s = 1|x, y) } P(x, y|s = 1)
\end{equation}

As initially proposed by Zadrozny \citep{zadrozny2004learning} there were four types of conditional dependencies:

\begin{itemize}
\item No \textit{sample selection bias} or $P(s = 1|x, y) = P(s = 1)$. In other words, the selection process is independent from both feature vector $x$ and class label $y$,
\item \textit{Feature bias} or $P(s = 1|x, y) = P(s = 1|x)$. The selection process is conditionally independent from $y$ given $x$ . It is important to understand that feature bias does not imply $s = 1$ is completely independent from $y$,
\item \textit{Class bias} or $P(s = 1|x, y) = P(s = 1|y)$. The selection process is conditionally independent from $x$ given $y$. Similarly, it does not imply that $s = 1$ is completely independent from $x$,
\item \textit{Complete bias} or $P(s = 1|x, y)$. The selection process is dependent on both $x$ and $y$.
\end{itemize}

There are many research studies relate to the SSB \citep{elkan2001foundations, zadrozny2003cost, zadrozny2001learning, zadrozny2004learning, fan2005improved, dudik2006correcting}. One approach for this problem is \textit{important sampling} that assign a weight to each data point \citep{zadrozny2003cost, zadrozny2004learning, fan2005improved}. The $P(x,y)$ re-writes in term of $P(x, y|s = 1)$ then we could do sampling with the weight $\dfrac{ P(s = 1) }{ P(s = 1|x, y) }$, but it is not an easy task due to estimating $P(s = 1|x, y)$ is not straightforward. Instead of trying to find what sample well represents the dataset, another simpler approach to handle SSB is using ensemble learning to combine results from multiple samples or from multiple classifiers \citep{fan2007sample}.

In the previous section, we preferred to use Random Forest with the undersampling method and this approach could lead to the Sample Selection Bias problem. In order to avoid it, we propose using ensemble method that will combine multiple Random Forests classifiers. Splitting the dataset into multiple subsets by applying the undersampling technique, each subset will be used as the training set of Random Forest. A final prediction could be mean or mode of these classifiers and it makes the final result more stable because we use all of the available data. We will give more details of this method in the following chapter.

\section{Feature Engineering}
\label{feature_engineering}

One of the most important steps in Machine learning is Feature Engineering which we will pre-process features in the dataset to improve the accuracy of a algorithm. The Feature Engineering usually consist: impute missing values, detect outliers, extract new useful features, \dots. A set of raw attributes in many credit card datasets is quite similar because the data collected from a transaction must comply with international financial reporting standards (American Institute of CPAs, 2011) \citep{aicpa}. Typical attributes in one credit card dataset are summarized in table \ref{tab:typical_raw_attributes}, these attributes could be grouped into four levels:

\begin{itemize}
\item Transaction level: these features typically are the raw attributes of the transaction which are collected in real time,
\item Card level: spending behavior of a card could be computed by aggregating information from transactions made in last given hours of a card,
\item User level: spending behavior of the customer, in some cases one user has only one card and these levels are the same,
\item Network level: users in our system are not isolated, they are connected and share the behavior, we will discuss this level in the next section.
\end{itemize}

\begin{table}
\centering
\footnotesize
\caption{List of typical raw attributes in a transaction dataset}
\label{tab:typical_raw_attributes}
\begin{center}
 \begin{tabular}{||c | c||} 
 \hline
 Attribute name & Description \\ [0.5ex] 
 \hline\hline
 Transaction ID & Transaction identification number \\ 
 \hline
 Time & Date and time of the transaction \\
 \hline
 Account number & Identification number of the customer \\
 \hline
 Card number & Identification of the credit card \\
 \hline
 Transaction type & ie. Internet, ATM, POS, ... \\
 \hline
Entry mode & ie. Chip and pin, magnetic stripe, ... \\
\hline
Amount & Amount of the transaction in Euros \\
\hline
Merchant code & Identification of the merchant type \\
\hline
Merchant group & Merchant group identification \\
\hline
Country & Country of transaction \\
\hline
Country 2 & Country of residence \\
\hline
Type of card & ie. Visa debit, Mastercard, American Express... \\
\hline
Gender & Gender of the card holder \\
\hline
Age & Card holder age \\
\hline
Bank Issuer & bank of the card \\
 [1ex] 
 \hline
\end{tabular}
\end{center}
\end{table}

There are many studies use only raw transactional features, e.g. time, amount, \dots and don't take into account the spending behavior of the customer, which is expected to help discover fraud patterns \citep{lebbe2008artificial}. Standard feature augmentation consists of computing variables such as average spending amount of the cardholder in the last week or last month, number of transactions in the same shop, number of daily transactions, \dots \citep{dal2014learned, krivko2010hybrid, whitrow2009transaction, bhattacharyya2011data, jha2012employing}.

Whitrow et al. proposed a transaction aggregation strategy in order to aggregate customer behavior \citep{whitrow2009transaction}. The computation of the aggregated features consists in grouping the transactions made during the last given number of hours by each categorical features, e.g. card id, user id, transaction type, merchant group, country or other, followed by calculating the average/total/\dots amount spent on those transactions. This methodology has been used by a number of studies \citep{bhattacharyya2011data, bahnsen2013cost, dal2014learned, jha2012employing, sahin2013cost, tasoulis2008mining, weston2008plastic}. Whitrow et al. \citep{weston2008plastic} proposed a fixed time frame to be 24, 60 or 168h, Bahnsen et al. \cite{bahnsen2016feature} extended time frames to: 1, 3, 6, 12, 18, 24, 72 and 168h.

The process of aggregating features consists in selecting those transactions that were made in the previous $t_p$ hours, for each transaction $i$ in the dataset $S$:

\begin{equation}
S_{agg} = T_{agg} (S, i, t_p) \\
\quad = \lbrace x_j^{amount} \mid ( x_j^{feature} = x_i^{feature} ) \wedge (H(x_i^{time}, x_j^{time}) < t_p) \rbrace
\end{equation}

where:

$T$: function creates a subset of S associated with a transaction $i$ with respect to the time frame $t_p$,

$x_i^{amount}$: the amount of transaction $i$,

$x_i^{feature}$: the categorical feature of transaction $i$ which is used to group the transactions,

$x_i^{time}$: the time of transaction $i$,

$H(t_1, t_2)$: number of hours between the times $t_1$ and $t_2$.

After that, we can compute a customer behavior in last given hours, for example, an average amount of this time frame:

\begin{equation}
x_i^{average\_amount} = \dfrac{1}{N} \sum_{x^{amount} \in S_{agg}} x^{amount}
\end{equation}

with:

$x_i^{average\_amount}$: a new feature computed from subset $S_{agg}$ by function average,

$N$: number of transactions in subset $S_{agg}$.

These new useful features above for capturing customer spending patterns, Bahnsen et al. \cite{bahnsen2016feature} are also interested in analyzing the time of a transaction. The logic behind this is a customer is expected to make transactions at similar hours. However dealing with the time of the transaction is not the same as dealing with the above features since the time in a day is the circle, therefore it is easy to make the mistake of using the arithmetic mean of transactions' time. They have proposed to overcome this limitation by modeling the time of the transaction as a periodic variable, in particular using the von Mises distribution \citep{fisher1995statistical}.

Recently Wedge et al. \citep{wedge2017solving} tried applying automated feature engineering with some simple functions to reduce the time cost of this time-consuming feature engineering step, they claimed that a number of false positives dropped by 54\% compared to their previous approach. In summary, the feature pre-processing is important and some above methods are simple, easy to compute and run in real time. At least, the fraud detection system should have some simple aggregations, e.g. average, for some periods, e.g. 24h.

\section{Measurement}
\label{measurement}

The most common measure for classification tasks is accuracy; however, in the imbalanced dataset it is a misleading assessment measure, as well as some other measures e.g. MME, BER, TPR and TNR, \dots. There are some measures could handle this problem, e.g. AUC, F-measure,\dots and a well-accepted measure for imbalanced classification is the Area Under the ROC Curve (AUC) \citep{chawla2009data}. AUC shows that how much the ROC curve is close to the perfect classification; however, Hand \citep{hand2009measuring} considers the standard calculation of the AUC as inappropriate because it making an average of different misclassification costs for classes. F-measure is more accurate in the sense that it is a function of a classifier and its threshold setting, i.e. it considers both the precision and the recall of the test to compute the score. The traditional F-measure or balanced $F_1$ score is the harmonic mean of precision and recall.

In many Fraud Detection System \citep{sahin2013cost, mahmoudi2015detecting, bahnsen2015example}, cost-based measures are defined to quantify the monetary loss due to fraud \citep{bahnsen2013cost} by computes average cost-matrix which is similar to the confusion matrix. Elkan \citep{elkan2001foundations} states that it is safer to assess the cost-sensitive problem in terms of benefit (inverse of cost) because there is the risk of applying different baselines when using a cost-matrix to measure overall cost. Dealing with this issue, normalized cost or savings \citep{bahnsen2015example} are used to judge the performance w.r.t. the maximum loss.

In the cost matrix, the cost of a missed fraud is often assumed to be equal to the transaction amount \citep{elkan2001foundations, bahnsen2013cost}, because it has to be refunded to a customer. Cost of false alert is considered to be equivalent to the cost of a phone call because our investigators have to make the phone call to the card-holder to verify a transaction whether it is a fraud or not. Furthermore, there are many intangible costs, e.g. reputation cost of a company, maintenance cost of the investigator's department, \dots. For all of these reasons, define a good cost measure is not easy in credit card fraud detection and there is no agreement on which is an appropriate way to measure the cost of frauds.

In a scenario with limited resources, e.g. there are only a few investigators, they can't check all alert transactions which are marked as fraudulent from the detection system. They have to put their effort into investigating transactions with those highest risk of fraud, in other words the detection system has to give each transaction its posterior fraud probability. With this requirement, the measure must contain the probability part in order to measures how good the system gives the high score to fraud transaction and reverse.

In summary, define a good measure to analyze the accuracy of the system is not easy, not only in our case but also in many other domains of Machine learning. Similar with the imbalanced problem, the cost-based measures seem very promising and we also need to know all of the cost for each instance. The $F-1$ score is better than all of the others, it is easy to compute and suitable for the imbalance problem, and we thereby will use the $F-1$ measure to evaluate our fraud detection pipeline.

\section{Fraud Network}
\label{fraud_network}

In the very early age of the computer, experts have noticed that a fraudulent account is often connected to another fraudster. This is true in many domains, for example in the telecommunication domain, therefore analyzing the links in the data could discover fraud networks and improve the accuracy of the system \citep{fast2007relational}.

The network not only could increase the true positives of our system, but also decrease the false positives. For example, consider the following legal scenario: a husband and his wife traveling to another country, the first transaction of the husband could create a fraud alert then the next transaction of his wife from the same city should be legal with or without confirmation of the first one.

Despite these advantages, aggregating fraud network-level features is not easy and recently had been taken with more consideration. Van Vlasselaer et al. have proposed APATE \citep{van2015apate}, a framework for credit card fraud detection that allows including network information as additional features describing a transaction. In the next year after APATE, they also proposed GOTCHA! \citep{van2016gotcha}, a novel approach which can improve the accuracy of traditional fraud detection tools in a social security context, this approach uses a time-weighted network and features extracted from a bipartite graph. They show that network-level features are able to improve significantly the performance of a standard supervised algorithm.

In summary, finding the network in the dataset is a complicated task and it's not the best choice for the first version of the fraud detection system.

\section{Concept Drift}
\label{concept_drift}

As mentioned in section \ref{sample_selection_bias}, the non-stationary environment could occur when we modify the dataset. In case a data generating source changes itself over time in an unforeseen manner, it is known as Concept Drift \citep{gama2014survey} or Dataset Shift \citep{quionero2009dataset}. With Bayes rule, we have a joint distribution of a sample $(x, y)$ is:

\begin{equation}
P(x, y) = P(y|x)P(x) = P(x|y)P(y)
\end{equation}

In classification tasks, we usually want to estimate the probability $P(y|x)$, from the above formula, we have:

\begin{equation}
P(y|x) = \frac{P(x|y)P(y)}{P(x)}
\end{equation}

From this Bayes formula, the change of data distribution at time $t$ and $t + 1$ could come from several sources\citep{kelly1999impact}:

\begin{itemize}
\item $P(y|x)$: a change occurs with a class boundary $( P_t(y|x) \neq P_{t+1}(y|x) )$, this makes any classifiers which are well designed will be biased,
\item $P(x|y)$: this change $( P_t(x|y) \neq P_{t+1}(x|y) )$ is a internal change within a class which is observed in an inside class but the class boundary isn't affected, also known as Covariate Shift \citep{moreno2012unifying},
\item $P(y)$: change in prior probability $( P_t(y) \neq P_{t+1}(y) )$, it makes a good design classifier become less reliable,
\item or combination of these three parts.
\end{itemize}

There is no $P(x)$ in the list because the change in $P(x)$ does not affect $y$ so we could ignore it \citep{hoens2012learning}.  In general, it is hard to say where the change comes from since we only see the generated data and we do not know or cannot control the data generating process. Learning in the non-stationary environment make us have to update models frequently to capture up-to-date patterns and also remove irrelevant patterns.

In fraud detection problem, we saw that fraudsters always try to create new fraudulent strategies to bypass our detection system, and new fraudsters also give a try with old strategies. Therefore, patterns in our system are learned from the past data may re-occur in the future then the removing process makes us losing the accuracy, this problem is known as the stability-plasticity dilemma \citep{grossberg1988nonlinear}.

In summary, dealing with Concept Drift problem, our fraud detection system has to update frequently to capture new fraudulent patterns as soon as possible, e.g. a bank has up-to-date labels at night then they could update their system daily.

\section{Delayed True Labels}
\label{delayed_true_labels}

Most of the supervised learning algorithms are based on the assumption that labels of the dataset are correct; however, in the real world environment it could be not true. In our case, the fraud detection problem, after the system generates alerts then with a limited number of investigators, only a restricted quantity of alerts could be checked. It means a small set of labeled transactions returned as \textit{feedback} and makes the unrealistic assumption that all transactions are correctly labeled by a supervisor.

Furthermore, non-alerted transactions are a large set of unsupervised instances that could be either fraudulent or genuine but we only know the actual label of them after customers have possibly reported unauthorized transactions and maybe available several days later. And the customers have different habits, e.g. rarely check a transcript of credit card given by the bank, which makes our dataset non-accurate and hard to modeling the fraudulent patterns.

In fraud detection problem, a company, e.g. a bank, usually has all latest up-to-date labels of transactions at night, and other companies could have after a longer time. It means that the system has to be updated frequently, i.e. as soon as possible after receives accurate labels. And the system which is never updated will lose their accuracy over the time.

Pozzolo et al. \citep{dal2015credit} claimed that the accurate up-to-date labels are very important and they proposed a learning strategy to deal with the feedback. Their method used both feedbacks $F_t$ and delayed supervised instances $D_{t-\delta}$:

\begin{itemize}
\item  a sliding window classifier $W_t$: daily updated classifier over the supervised samples received in the last $\delta$ + M days, i.e. $ \lbrace F_t, \dots, F_{t - (\delta - 1)}, D_{t-\delta}, \dots, D_{t- (\delta + M -1)} \rbrace $,
\item an ensemble of classifiers $\lbrace \mathcal{M}_1, \mathcal{M}_2, \dots, \mathcal{M}_M, \mathcal{F} \rbrace$ where $\mathcal{M}_i$ is trained on $D_{t - ( \delta + i - 1)}$ and $F_t$ is trained on all the feedbacks of the last $\delta$ days $\lbrace F_t, \dots, F_{t - (\delta - 1)} \rbrace$.
\end{itemize}

Their solutions used two basic approaches for handling concept drift that can be further improved by adopting dynamic sliding windows or adaptive ensemble sizes \citep{fan2004systematic}. In summary, to keep the accuracy of the system, we need to update all of the classifiers as soon as the data available, and usually it is a daily update.

\section{Performance}
\label{performance}

Particularly in fraud detection problem, we are dealing with a huge credit card transaction dataset. As we see in the sections above, many approaches are used to detect fraud transactions, many of them have shown that they have a good result with their approach, however, most of them could not be used in production because they did not consider their running time. For instance, in the age of Neural Network there are many studies try to use Neural Network models to detect fraud, but this approach runs the training process very slowly. 

In the age of Big Data, there are some frameworks for distributed computing system and the most common one currently used in many biggest system around the world is Spark \cite{zaharia2010spark}. Spark is an open-source cluster-computing framework, originally developed at the University of California, Berkeley's AMPLab. With Spark framework, we could distribute the computation for each transaction into one node in our cluster, and the cluster is easy to scale up.

Spark also supports Machine learning algorithms via its MLlib library \cite{meng2016mllib}. However, with the current version of Spark (2.2.0 \cite{spark220}), there are not many algorithms supported, which means that for those companies they have only a few options to build the detection system. Fortunately, some of the most common algorithms are supported, e.g. Linear Model, Decision Tree or Random Forest, \dots.

\section{Fraud Detection Pipeline}

We have discussed the challenges of building the Fraud Detection System and found out several possible ways to resolve them. Based on that, in this section, we analyze and propose a Fraud Detection Pipeline, which is not only easy to implement but also has a reliable result.

\subsection*{Imbalanced dataset and Algorithm}

The imbalance problem and algorithms in the fraud detection problem could be grouped into two groups based on our point of view:

\begin{itemize}
\item Theoretical perspective: apply Data Science and Data Mining to obtain a \textit{good} result like other applications,
\item Financial perspective: this task is a problem in financial institutions and they want to minimize their losing money.
\end{itemize}

However the detail of losing money is usually not published since it is sensitive data, therefore those cost-based methods are out of the scope of this report. Among the rest approaches, Pozzolo et al. \citep{dal2013racing} have shown that the Random Forest combined with undersampling or SMOTE often has the best result. In our case, we prefer undersampling over SMOTE because it could manage the huge dataset. Usage details of the undersampling and Random Forest algorithm are provided later.

\subsection*{Sample Selection Bias}

\begin{figure}
\centering
\caption{Building single model by combining multiple Random Forest classifiers (RF)}
\label{img:single_rf}
\includegraphics[scale=0.7]{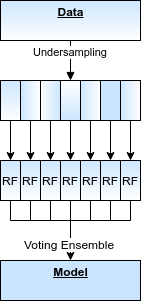}
\end{figure}

A chosen sampling technique, undersampling, leads to the Sample Selection Bias problem and in order to avoid it, we propose to use an ensemble method that combines results from multiple samples \citep{fan2007sample}. The dataset will be split into multiple dis-joint samples, each of which is used to train a Random Forest classifier then the final prediction could be the mean or the mode value of these classifiers (see figure \ref{img:single_rf}). This approach is also useful for the imbalance problem above and reduces training times of the Random Forest classifiers.

\subsection*{Feature Engineering and Fraud Network}

Feature Engineering is one of the most important steps in Data Science and usually is difficult, time-consuming, and requires expert knowledge. We don't discuss in a detail this step, but in summary features after this step could be grouped into four levels:

\begin{itemize}
\item Transaction level: typically is the raw attributes of a transaction, which are available in real-time,
\item Card level: this level describes a spending behavior of a credit card, which is computed by aggregating the features from its previous transactions,
\item User level: usually same as card level if one user has only one card, it describes the spending behavior of a customer and a computing method is the same as card level,
\item Network level: detect Fraud Network in our system could improve the accuracy significantly but it is a complicated task, therefore we skip this level in our pipeline.
\end{itemize}

We propose to aggregate features which describe customer's behavior (user level and/or card level) consisting grouping transactions made during a last given number of hours by each categorical features, e.g. card id, user id, transaction type, \dots; followed by calculating some simple functions, e.g. average, amount spent on those transactions. The time frames could be 1, 3, 6, 12, 18, 24, 72, 168 hours and more. The detailed formulas are in section \ref{feature_engineering}.

\subsection*{Measurement}

Some common measures could not be used to evaluate the Fraud Detection System due to the imbalance problem. From the theoretical perspective, the simplest and suitable measure to deal with this is F1-score. Or inside a financial company, with all cost information on each transaction, we could use cost-based measures for those cost-based methods. In this thesis, we only use F1-score for its simplicity.

\subsection*{Concept Drift}

The data, which is fraudulent patterns and customer's behavior, always change over time. Dealing with this Concept Drift problem, all classifiers in our detection system have to update frequently to capture new patterns as soon as possible. We assume that we could update the system daily or at least every time frame, which is possible with a small dataset, and in the remaining of this chapter we will update all models at all time frames.

\subsection*{Performance}

As we have described in section \ref{performance} that to bring this Fraud Detection Pipeline into production, we have to consider its performance. And fortunately for us, the distributed computing framework Spark could handle our requirements, e.g. distribute our computation or run Machine Learning algorithms parallel, \dots. Therefore we propose to use this framework in both the development and the production phase.

\subsection*{Fraud Detection Pipeline}

With the Fraud Detection Pipeline architecture, we propose to build several models on different time frames for multiple purposes, such as:

\begin{itemize}
\item short-term model: using small latest data, e.g. one time frame data, to captures newest fraudulent strategies. In real life, it could be the last day or the last week data,
\item intermediate-term model: using more data than the above, i.e. longer period than short-term, e.g. two time frame data, to capture not only newest but also older fraudulent strategies. In real life, it could be the last month or last year data,
\item long-term model: similar with the above models but using more or all of the available data, e.g. three time frame data, to keep all possible fraudulent strategies in our data.
\end{itemize}

\begin{figure}
\centering
\includegraphics[width=\textwidth]{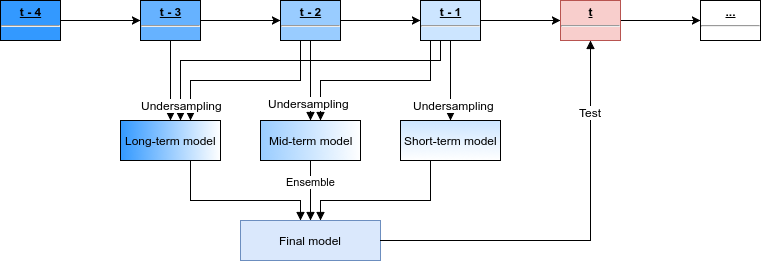}
\caption{Recommended Fraud Detection Pipeline}
\label{img:fraud_detection_pipeline}
\end{figure}

The pipeline is summarized in figure \ref{img:fraud_detection_pipeline} and the detail of each model have been described above (see figure \ref{img:single_rf}). The size of the data and its training time increase from short-term to long-term model to accumulate fraudulent patterns. The long-term model needs a long time to prepare its new version and could be not available for the next day data, but with the fastest model, we could detect new fraudulent patterns while the new long-term model is training. This architecture could reduce workload for the slowest model and makes sure the system is always prepared for any possible situation.

\section{Upgrade the Fraud Detection Pipeline}

Since our proposed Fraud Detection Pipeline is a simple approach, there are several things could be done to improve the system and here we summarize some of the most potential improvements that will upgrade the current system easily:

\begin{itemize}
\item Imbalanced dataset: there are two main ideas to resolve this problem, one is the sampling methods and the other is cost-based methods. Our pipeline using the simpler way that does not need to know about the cost but it is very promising for the business view. We suggest trying to use the cost-based methods to have an overview of loss or saving money while using the system,
\item Algorithm: the supervised Random Forest algorithm had chosen as the main classifiers in the pipeline. We suggest expanding the models in the pipeline that could use more data or use other Machine Learning algorithms, e.g. the unsupervised learning to detect anomaly transactions or use stronger algorithms like Neural Networks, then finally combine all of classifiers to make the final prediction,
\item Feature Engineering: there are some profiling methods to capture customer's behaviors as we have described in section \ref{feature_engineering}. While the original attributes in the credit card dataset is limited, these aggregated features are very useful for the system and we strongly propose to apply all of these methods as the simplest way to upgrade the pipeline,
\item Measurement: the F1 score is the most suitable choice for the imbalance problem, therefore we do not need to change the metric. However, if we have the cost of each transaction, with or without the cost-based methods for both imbalance problem and algorithms, we suggest having one cost-based measure to see how much the money the system could save for a company,
\item Fraud Network: finding a network of fraudulent transactions in the data could improve the accuracy significantly. However, this task is not easy and requires complex researches so we suggest the readers only give a try with it after resolving all of the other things,
\item Delayed True Labels: the delayed labels are very important since it is the up-to-date and accurate labels. We suggest extending the Algorithms section above with one model using this data and with wrong predictions of our models, it could not only predicts the true delayed labels but also explains why our prediction fails,
\item Performance: the FDS usually need to process a very large number of transactions. While the distributed-computing Spark framework could handle it and also fits with other industry's requirements, therefore we do not have any reason to find an alternative solution and we suggest using the Spark framework in our system.
\end{itemize}

\section{Experiments and results}

Credit card transactions are very sensitive data, therefore there are only a few public datasets on the internet, some of them are very obsolete and do not have too many transactions. Fortunately there is one new and large transaction dataset which is published by Pozzolo et al. \citep{dal2015calibrating} in 2016, this dataset contains transactions made in September 2013 by European cardholders. It presents transactions that occurred in two days, where we have 492 frauds out of 284,807 transactions. This dataset is highly unbalanced, the positive class (fraudulent transactions) is 0.172\% of all transactions, i.e. imbalance ratio is 577.

The dataset contains only numerical input variables which are the result of a PCA transformation. Unfortunately, due to confidentiality issues, they cannot provide the original features and more background information about the data. Features \textit{V1}, \textit{V2},\dots, \textit{V28} are the principal components obtained with PCA; the only features which have not been transformed with PCA are \textit{Time} and \textit{Amount}. The feature \textit{Time} contains the seconds elapsed between each transaction and the first transaction in the dataset, the feature \textit{Amount} is the transaction's amount. Column \textit{Class} is the response variable and it takes value 1 in case of fraud and 0 otherwise.

\begin{figure}
\centering
\includegraphics[width=\textwidth]{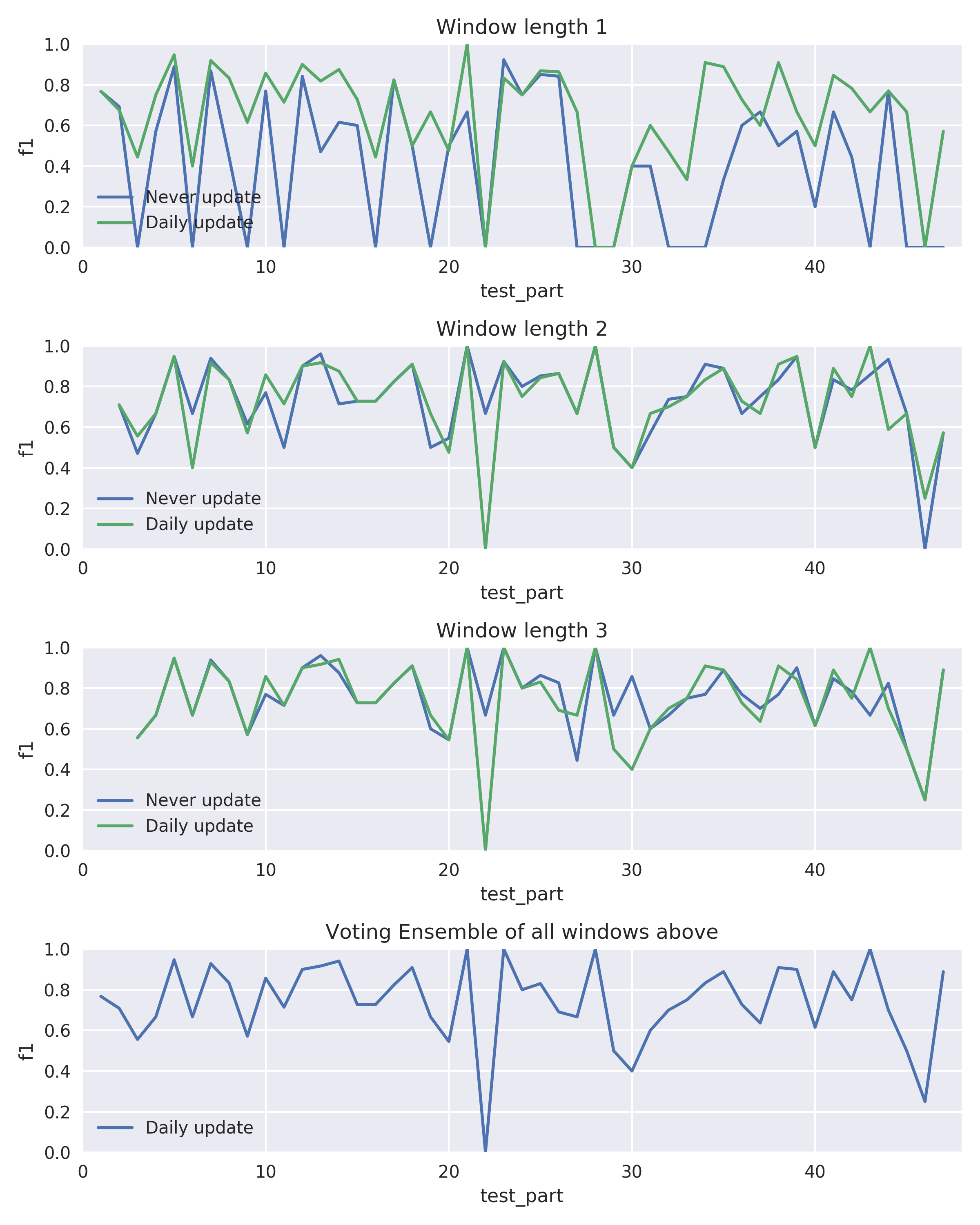}
\caption{Accuracy of our Fraud Detection Pipeline}
\label{img:fdp_summary}
\end{figure}

In this thesis, we use this dataset to test the proposed Fraud Detection Pipeline and also for other studies below. Splitting this dataset into 48 time frames, we use the first 24 time frames for a training phase and the rest for the testing phase. Testing with three window lengths, short-term, intermediate-term and long-term periods are 1,2 and 3 time frames respectively. We run a stratified 5-fold cross-validation then compute the metric $F_1$ score as an average of five folds. This study runs on one single machine with 24 cores CPU and 128 gigabytes memory. The result is in figure \ref{img:fdp_summary} and a detail is in table \ref{tbl:fdp_summary}. 

In the result, the short-term model often fails with zero $F_1$ score if it does not re-train anymore and compare with daily update, the model will be improved significantly (81\%). Only the long-term model with the daily update has a bit lower accuracy than the never update model (1\%). As we forecast above, the more data we use in the longer term model the more accuracy we have, i.e. the long-term model with $F_1$ score is 22\% higher than the short-term model. In general, the daily update mechanism usually better in both average and standard deviation of $F_1$ scores over 5-fold cross-validation.

\begin{table}[]
\centering
\footnotesize
\caption{$F_1$ results of the Fraud Detection Pipeline}
\label{tbl:fdp_summary}
\begin{tabular}{|c|c|c|c|c|c|c|}
\hline
\multirow{2}{*}{Window length} & \multicolumn{2}{c|}{Update mechanism}                                    & \multicolumn{2}{c|}{Training $F_1$} & \multicolumn{2}{c|}{Testing $F_1$} \\ \cline{2-7} 
                               & Strategy     & \begin{tabular}[c]{@{}c@{}}Time frame\\ (on testing set)\end{tabular} & Average           & Std             & Average              & Std         \\ \hline
\multirow{2}{*}{1}             & Never update & 0                                                         & 0.4758            & 0.3480          & 0.3331               & 0.3242      \\ \cline{2-7} 
                               & Daily update & 24                                                        & 0.6951            & 0.2312          & \textbf{0.6024}      & 0.2808      \\ \hline
\multirow{2}{*}{2}             & Never update & 0                                                         & 0.7506            & 0.1632          & 0.72                 & 0.2192      \\ \cline{2-7} 
                               & Daily update & 24                                                        & 0.7325            & 0.2304          & \textbf{0.722}       & 0.1893      \\ \hline
\multirow{2}{*}{3}             & Never update & 0                                                         & 0.7808            & 0.1511          & \textbf{0.7351}      & 0.1665      \\ \cline{2-7} 
                               & Daily update & 24                                                        & 0.7571            & 0.2267          & 0.7268               & 0.1871      \\ \hline
Ensemble                & Daily update & 72                                                        & 0.7554            & 0.2164          & \textbf{0.7261}      & 0.1865      \\ \hline
\end{tabular}
\end{table}

\section{Conclusion and other works}

In this survey, we have discussed about several challenges in building an effective Fraud Detection System and then proposed one most suitable and also easy-to-use approach to build our Fraud Detection Pipeline. From the $F_1$ score in the results, it shows that our pipeline could handle the fraud detection problem effectively only with the simplified approach. We also suggest for the readers many things to do to upgrade our pipeline, and it leads to three independent works below:

\begin{itemize}
\item Concept Drift: we have assumed that we could update all of the models in our system every day; in case the data is very huge and the update process takes more than one day to complete then it is a costly task for our system. In chapter 4, we propose a method to decide which model needs to update with the most up-to-date data and we call it as Update Point Estimation algorithm,
\item Imbalance problem: in our proposed pipeline, we used undersampling technique and ensemble learning to handle the imbalance problem (see figure \ref{img:single_rf}) in the credit card dataset which is not only huge but also imbalanced extremely. In chapter 5, we show that why this combination is a most suitable in our case,
\item Algorithms: we only use the Random Forest algorithm in our pipeline, in the rich of literature in fraud detection problem almost algorithms, e.g. supervised learning and unsupervised learning, are used. To the best of our knowledge, a graph-based semi-supervised learning techniques have not been applied to this problem. Therefore in chapter 6, we propose to use an un-normalized graph p-Laplacian based semi-supervised learning combined with the undersampling technique to the fraud detection problem.
\end{itemize}

\chapter{Update Point Estimation for the case of Concept-Drift in fraud detection problem}

\ifpdf
    \graphicspath{{Chapter4/Figs/Raster/}{Chapter4/Figs/PDF/}{Chapter4/Figs/}}
\else
    \graphicspath{{Chapter4/Figs/Vector/}{Chapter4/Figs/}}
\fi

In the previous section, we proposed the Fraud Detection Pipeline and to avoid the Concept Drift problem, i.e. to keep its accuracy, it needs to update frequently, e.g. daily. In fact, it only needs to update to adapt to new patterns in the data. The Concept Drift is usually seen in streaming data and several methods have been proposed to deal with it, but in the fraud detection problem the data is often the batch data and these methods are not suitable for this case. We propose a method to detect when we should update our models and the result shows that our models not only need to update less frequently but also have the better performance.

\section{Introduction}

Imbalance is one common issue in many real domains, e.g. fraud detection, telecommunication. In those cases, we need to identify a small number of positive data points (minority class) stand among too many redundant data points. Consider a classification task of a dataset with imbalance ratio (IR) of 100, i.e. in every 101 samples there is only one positive sample that we need to detect. Most of algorithms in Machine Learning are not designed to handle this situation; if they maximize their accuracy then in the worst case they always have the accuracy of 99\% by doing nothing. This lazy classifier marking all samples in a dataset as majority class has very high accuracy, but mis-classify all minority samples. Many studies have reported that they lose their performance with the imbalance datasets. \cite{chen2004using,chen2005pruning,wang2006classification,hong2007kernel}.

To handle the imbalance problem, there are many proposed methods and these could be grouped into two levels: algorithmic level and data level. At the algorithmic level, classifiers are designed or modified from existing algorithms to handle the imbalanced data by itself \cite{bradford1998pruning,cieslak2008learning}. At the data level, original imbalanced data will be modified by pre-processing step before applying to normal classification algorithms, e.g. under-sampling, over-sampling \cite{drummond2003c4} or SMOTE \cite{chawla2002smote}. Several studies \cite{weiss2001effect, laurikkala2001improving, estabrooks2004multiple} have shown that a balanced training set will give the better performance when using normal algorithms. Cost-sensitive approach \cite{ling2004decision} uses a similar idea with those above approaches to minimize misclassification costs, which are higher for data points in the minority class; however, it requires side information about the cost which is not always available.

In the age of computing world, the size of data quickly increases to a huge volume due to the advantages of computer technologies. Big dataset could be seen in many domains like genome biology, banking system. The size of the dataset could be million or billion records, which leads to the question that whether our solution effective with a massive amount of data. Particularly in an imbalanced big data classification \cite{del2014use, triguero2015evolutionary, fernandez2017insight}, previous approaches could not provide solutions since they create a bigger dataset then cannot fit into our resources or has poor results. Especially, an extremely imbalanced data problem in a big dataset is another story that receives attention very recently \cite{triguero2015rosefw, krawczyk2016learning}.

Several studies in imbalanced datasets have imbalance ratio less than 100, i.e. minority class greater than 1\% of the data, and those approaches cannot guarantee the accuracy for the highly imbalanced tasks (IR $>$ 100). In some applications, the datasets are not only really huge but also highly imbalanced, e.g. fraud detection often has IR greater than 1000 \cite{juszczak2008off}. From this scenario of two difficult problems, we need an effective way to tackle both of them at the same time and therefore, in this study we want to show a simplified combination of under-sampling technique and ensemble learning could have a good result in the extremely imbalanced big data classification.

The paper is organized as follows: section 2 mentions recent works relate to the extremely imbalanced data, imbalance big data classification and the gap between them. Section 3 presents our methodology, experimental design and comparative result on many datasets. Then we finally come to a conclusion and future works in section 4.

\section{Recent works on extreme imbalance in big data classification}

In the class imbalance problem, Mikel Galar at el. \cite{galar2012review} gave a comprehensive review of several methodologies that have been proposed to deal with this problem. From their empirical comparison of the most significant published approaches, they have concluded that ensemble-based algorithms, e.g. using bagging or boosting, are worthwhile and under-sampling techniques, such as RUSBoost or UnderBagging,  could have higher performances than many other complex approaches.

In the Big Data issue, techniques used to deal with it usually based on distributed computing on a system of computers. Similarly in order to handle the imbalance problem in the big dataset, Sara del Río at el. \cite{del2014use} implemented a distributed version of those common sampling techniques based on MapReduce framework \cite{miner2012mapreduce} and they found that the under-sampling method could manage large datasets. Isaac Triguero at el. \cite{triguero2015rosefw} faced with the extremely imbalanced big data in bioinformatics problem and their solution is a complex combination of Random Forest \cite{breiman2001random} and oversampling that balanced the distribution of classes. In a review of Alberto Fernández at el. \cite{fernandez2017insight} on the Big Data and Imbalanced classification, they considered different sampling ratio between classes and stated that a perfectly balanced sample was not the best method as traditional approaches that have been used \cite{hido2009roughly, garcia2009evolutionary}.

Recently, some researchers notice the extremely imbalance problem on datasets with imbalance ratio greater than 100 \cite{tang2009svms, triguero2015rosefw}. This context easily leads to a classification task in the Big Data since the positive class is a very small fraction in the dataset, if the dataset is small then we cannot find any meaningful patterns on a few positive data points. To our best knowledge, there are a few studies on this gap of extremely imbalanced big data classification. For example, Sara del Río \cite{del2014use} tested their study on datasets with a largest size is over five million data points or a maximum IR is 74680. This new scenario raises a question about how effectiveness we could achieve with this kind of data. An ideal method need to use any given data points very effective, for both positive and negative data points.

In the traditional imbalance problem, many evaluation measures could not be used in this case independently since they are affected by the majority class. In order to evaluate the classifier, it requires a stronger metric that balance the classes and a commonly used metric for this case is $F_1$ score that can derive from confusion matrix (table \ref{tbl:confusion_matrix}). Based on two other metrics, Precision and Recall, the $F_1$ score is the harmonic average of them and could be computed as formula \ref{eq:f1}. In this study we propose to use the $F_1$ score as a main measure.

\begin{table}[h]
\centering
\caption{Confusion matrix}
\label{tbl:confusion_matrix}
\begin{tabular}{lll}
\hline
               & predicted positives & predicted negatives \\ \hline
real positives & True positive (TP)  & False negative (FN) \\ \hline
real negatives & False positive (FP) & True negative (TN)  \\ \hline
\end{tabular}
\end{table}

\begin{equation}
\label{eq:precision}
Precision = \dfrac{TP}{TP + FP}
\end{equation}

\begin{equation}
\label{eq:recall}
Recall = \dfrac{TP}{TP + FN}
\end{equation}

\begin{equation}
\label{eq:f1}
F_1 = 2 \cdot \dfrac{Precision \cdot Recall}{Precision + Recall}
\end{equation}

\section{Proposed analysis}

Based on the question posed in the previous section, we propose another way to build samples that effectively use all of the data points in the dataset once time. Using the under-sampling method, all negative data points will be randomly divided into several dis-joint same-size subsets then each subset will be merged back with all positive data points then it is used to train a classifier. After that, we combine these classifiers by the voting ensemble as a final prediction. Suppose a dataset $D$ has $N$ data points with imbalance ratio is $IR$, let $K$ be a number of subsets, our proposed approach is summarized by Algorithm \ref{alg:ksus}, in which we define $D^{major}$ and $D^{minor}$ are the majority and minority classes in the dataset $D$ respectively and $n = \begin{bmatrix} \dfrac{ \begin{vmatrix} D^{major} \end{vmatrix} }{ K } \end{bmatrix}$ is a size of a sampled subset. In the main part of Algorithm \ref{alg:ksus},  in this study we mainly use Random Forest because of its robustness and good performance.

\begin{algorithm}[h]
\caption{$K$-Segments Under Bagging ($K$-SUB)}
\label{alg:ksus}

Inputs:
\newline
$D^{major}$ is the majority class in the dataset $D$
\newline
$D^{minor}$ is the minority class in the dataset $D$
\newline
$K$ is the number of subsets

Procedure:
\begin{algorithmic}[1]
\For {$k = 1$ to $K$}:
  \State Draw a subset $K^{major}_k$ without replacement from the majority class with the size $n = \begin{bmatrix} \dfrac{ \begin{vmatrix} D^{major} \end{vmatrix} }{ K } \end{bmatrix}$
  \State Let $K_k$ is a merged subset of $K^{major}_k$ with all data points in minority class $D^{minor}$
  \State Train a classifier $f^k$ by applying Random Forest algorithm to the subset $K_k$
\EndFor
\State Combine all $f_k$ into an aggregated model $F$
\State Return F
\end{algorithmic}


\end{algorithm}


To illustrate the effectiveness of our proposed algorithm, we apply it on many datasets with various aspects, e.g. from small to large datasets, from low to very high imbalance ratio and moreover we try to do our test with several values of $K$, from 3, 5, 10 to 20. We run a stratified 5-fold cross-validation then compute the metric $F_1$ score as an average of five folds and the running time is also reported in seconds. This test is carried out in one single machine with 24 cores and 128 gigabytes in memory. All details of the final results are summarized in the table \ref{tbl:summary_results}.


\begin{table*}[]
\caption{Summary of datasets and the experimental results}
\label{tbl:summary_results}
\resizebox{\textwidth}{!}{%
\begin{tabular}{|c|c|c|c|c|c|c|c|c|c|c|c|c|c|}
\hline
\multirow{2}{*}{dataset} & \multirow{2}{*}{IR} & \multirow{2}{*}{sample} & \multirow{2}{*}{feature} & \multicolumn{2}{c|}{UB}  & \multicolumn{2}{c|}{3-SUB} & \multicolumn{2}{c|}{5-SUB} & \multicolumn{2}{c|}{10-SUB} & \multicolumn{2}{c|}{20-SUB} \\ \cline{5-14} 
                         &                     &                         &                          & F1              & time   & F1                & time   & F1                & time   & F1                 & time   & F1            & time        \\ \hline
ecoli                    & 8                   & 336                     & 7                        & \textbf{0.5758} & 1      & 0.5669            & 1      & 0.4850            & 2      & 0.4563             & 5      & 0.3953        & 9           \\ \hline
spectrometer             & 10                  & 531                     & 93                       & 0.7642          & 1      & \textbf{0.8411}   & 1      & 0.8248            & 2      & 0.7933             & 4      & 0.6601        & 10          \\ \hline
car\_eval\_34            & 11                  & 1728                    & 21                       & \textbf{0.7173} & 2      & 0.4445            & 1      & 0.2418            & 2      & 0.2380             & 5      & 0.2081        & 9           \\ \hline
isolet                   & 11                  & 7797                    & 617                      & 0.6084          & 5      & \textbf{0.7702}   & 3      & 0.7326            & 4      & 0.5984             & 7      & 0.4710        & 13          \\ \hline
libras\_move             & 14                  & 360                     & 90                       & 0.6797          & 2      & \textbf{0.7389}   & 1      & 0.6473            & 2      & 0.4036             & 4      & 0.1717        & 9           \\ \hline
thyroid\_sick            & 15                  & 3772                    & 52                       & 0.7551          & 2      & 0.8023            & 2      & \textbf{0.8261}   & 3      & 0.8112             & 5      & 0.6708        & 10          \\ \hline
oil                      & 21                  & 937                     & 49                       & \textbf{0.3083} & 3      & 0.0941            & 1      & 0.0829            & 2      & 0.0848             & 4      & 0.0841        & 9           \\ \hline
car\_eval\_4             & 25                  & 1728                    & 21                       & \textbf{0.6078} & 4      & 0.4413            & 1      & 0.2183            & 2      & 0.1623             & 5      & 0.1481        & 10          \\ \hline
letter\_img              & 26                  & 20000                   & 16                       & 0.6133          & 5      & \textbf{0.8681}   & 3      & 0.8381            & 4      & 0.8059             & 9      & 0.7018        & 15          \\ \hline
webpage                  & 34                  & 344780                  & 300                      & 0.3783          & 14     & 0.1842            & 5      & 0.3130            & 6      & \textbf{0.4496}    & 12     & 0.1616        & 18          \\ \hline
mammography              & 42                  & 11183                   & 6                        & 0.4033          & 5      & 0.6478            & 1      & \textbf{0.6606}   & 1      & 0.6316             & 3      & 0.5221        & 5           \\ \hline
protein\_homo            & 111                 & 145751                  & 74                       & 0.4877          & 24     & 0.7908            & 7      & 0.7953            & 7      & \textbf{0.8125}    & 12     & 0.8113        & 15          \\ \hline
10\% kddcup R2L          & 443                 & 494021                  & 41                       & 0.2396          & 643    & 0.8945            & 44     & \textbf{0.9350}   & 46     & 0.8803             & 90     & 0.7176        & 101         \\ \hline
fraud\_detection         & 577                 & 284807                  & 29                       & 0.2341          & 148    & \textbf{0.8113}   & 30     & 0.6231            & 30     & 0.5567             & 55     & 0.3304        & 62          \\ \hline
kdd SF PROBE             & 7988                & 703067                  & 30                       & 0.0126          & 3980   & 0.7895            & 80     & 0.8034            & 79     & \textbf{0.8242}    & 156    & 0.7176        & 156         \\ \hline
10\% kdd U2R             & 9499                & 494021                  & 41                       & 0.0075          & 7008   & \textbf{0.6034}   & 38     & \textbf{0.6034}   & 39     & 0.5782             & 79     & 0.5291        & 90          \\ \hline
kdd U2R                  & 94199               & 4940219                 & 41                       & 0.0007          & 112285 & 0.3741            & 359    & 0.4571            & 343    & \textbf{0.5617}    & 737    & 0.4916        & 780         \\ \hline
\end{tabular}%
}
\end{table*}

The results in bold indicate the best $F_1$ scores from the Under Bagging (UB) algorithm and our method $K$-Segments Under Bagging ($K$-SUB) with multiple $K$ values. With datasets have low imbalance ratio (e.g. IR $<$ 50) the Under Bagging algorithm could handle the problem, and our method has also comparative results. However, from a high to extreme imbalance cases, i.e. IR $>=$ 50, our method outperform the Under Bagging in all cases with any value of our chosen $K$ values. Furthermore, the running time is dramatically slow for the case of Under Bagging algorithm, but with our method it only takes us few minutes in the same machine. Specially in the two biggest datasets with highest $IR$, the $F_1$ scores of the Under Bagging tend to become zero but our method still keeps the high $F_1$ scores. In the range of $K$ values selected above, we observe the case $K = 20$ is not the best one, so it suggests that only a small enough value of $K$ may be a good choice in our proposed analysis.

\section{Conclusion}

In this study, we have presented the $K$-Segments Under Bagging algorithm in the attempt to tackle the problem of extremely imbalanced data classification. The experimental results show that in the case of extremely imbalanced data, our method not only outperforms the previous method but also runs very fast. While in case of low imbalanced data, with the suitable $K$ value, the $K$-SUB algorithm is still useful. Furthermore, it can be observed experimentally that we may obtain the good results for the small enough values of $K$, which suggests that in real applications we only have to tune the parameter $K$ in a small range of value. With this new and challenging scenario, we have shown that the simplified combination of undersampling technique and ensemble learning is able to give better results instead of using other complicated methods addressed in the past. In the future work, we want to investigate deeper on this approach as well as other related issues.

\chapter{K-Segments Under Bagging approach: An experimental Study on Extremely Imbalanced Data Classification}

\ifpdf
    \graphicspath{{Chapter1/Figs/Raster/}{Chapter1/Figs/PDF/}{Chapter1/Figs/}}
\else
    \graphicspath{{Chapter1/Figs/Vector/}{Chapter1/Figs/}}
\fi

Imbalanced dataset could be found in many real-world domains of applications, e.g. fraud detection problem, threat detection, etc.,. There are various methods that have been proposed to handle the imbalanced data classification problems, but there is no guarantee those methods will work well in the case of extremely imbalanced data. Practically, we can find the explosion of imbalance issue in the case of big dataset analysis, in which the imbalance ratio increases uncontrollably. We all know that Big Data is a terminology used to express the increasing level of both volume and complexity of the data, and the big data within extreme imbalance scenario is such a research question from recent work. In this study, we propose a simplified combination of under-sampling and ensemble learning which can adapt well with different scenarios of extreme imbalance. Experimentally, we carry out our test on 17 datasets, taken from the UCI repository and Kaggle, and can show that our proposed method is not only competitive with a common method but also very effective especially in the case of extremely imbalanced big data classification problems.

\section{Introduction}

Imbalance is one common issue in many real domains, e.g. fraud detection, telecommunication. In those cases, we need to identify a small number of positive data points (minority class) stand among too many redundant data points. Consider a classification task of a dataset with imbalance ratio (IR) of 100, i.e. in every 101 samples there is only one positive sample that we need to detect. Most of algorithms in Machine Learning are not designed to handle this situation; if they maximize their accuracy then in the worst case they always have the accuracy of 99\% by doing nothing. This lazy classifier marking all samples in a dataset as majority class has very high accuracy, but mis-classify all minority samples. Many studies have reported that they lose their performance with the imbalance datasets. \cite{chen2004using,chen2005pruning,wang2006classification,hong2007kernel}.

To handle the imbalance problem, there are many proposed methods and these could be grouped into two levels: algorithmic level and data level. At the algorithmic level, classifiers are designed or modified from existing algorithms to handle the imbalanced data by itself \cite{bradford1998pruning,cieslak2008learning}. At the data level, original imbalanced data will be modified by pre-processing step before applying to normal classification algorithms, e.g. under-sampling, over-sampling \cite{drummond2003c4} or SMOTE \cite{chawla2002smote}. Several studies \cite{weiss2001effect, laurikkala2001improving, estabrooks2004multiple} have shown that a balanced training set will give the better performance when using normal algorithms. Cost-sensitive approach \cite{ling2004decision} uses a similar idea with those above approaches to minimize misclassification costs, which are higher for data points in the minority class; however, it requires side information about the cost which is not always available.

In the age of computing world, the size of data quickly increases to a huge volume due to the advantages of computer technologies. Big dataset could be seen in many domains like genome biology, banking system. The size of the dataset could be million or billion records, which leads to the question that whether our solution effective with a massive amount of data. Particularly in an imbalanced big data classification \cite{del2014use, triguero2015evolutionary, fernandez2017insight}, previous approaches could not provide solutions since they create a bigger dataset then cannot fit into our resources or has poor results. Especially, an extremely imbalanced data problem in a big dataset is another story that receives attention very recently \cite{triguero2015rosefw, krawczyk2016learning}.

Several studies in imbalanced datasets have imbalance ratio less than 100, i.e. minority class greater than 1\% of the data, and those approaches cannot guarantee the accuracy for the highly imbalanced tasks (IR $>$ 100). In some applications, the datasets are not only really huge but also highly imbalanced, e.g. fraud detection often has IR greater than 1000 \cite{juszczak2008off}. From this scenario of two difficult problems, we need an effective way to tackle both of them at the same time and therefore, in this study we want to show a simplified combination of under-sampling technique and ensemble learning could have a good result in the extremely imbalanced big data classification.

The paper is organized as follows: section 2 mentions recent works relate to the extremely imbalanced data, imbalance big data classification and the gap between them. Section 3 presents our methodology, experimental design and comparative result on many datasets. Then we finally come to a conclusion and future works in section 4.

\section{Recent works on extreme imbalance in big data classification}

In the class imbalance problem, Mikel Galar at el. \cite{galar2012review} gave a comprehensive review of several methodologies that have been proposed to deal with this problem. From their empirical comparison of the most significant published approaches, they have concluded that ensemble-based algorithms, e.g. using bagging or boosting, are worthwhile and under-sampling techniques, such as RUSBoost or UnderBagging,  could have higher performances than many other complex approaches.

In the Big Data issue, techniques used to deal with it usually based on distributed computing on a system of computers. Similarly in order to handle the imbalance problem in the big dataset, Sara del Río at el. \cite{del2014use} implemented a distributed version of those common sampling techniques based on MapReduce framework \cite{miner2012mapreduce} and they found that the under-sampling method could manage large datasets. Isaac Triguero at el. \cite{triguero2015rosefw} faced with the extremely imbalanced big data in bioinformatics problem and their solution is a complex combination of Random Forest \cite{breiman2001random} and oversampling that balanced the distribution of classes. In a review of Alberto Fernández at el. \cite{fernandez2017insight} on the Big Data and Imbalanced classification, they considered different sampling ratio between classes and stated that a perfectly balanced sample was not the best method as traditional approaches that have been used \cite{hido2009roughly, garcia2009evolutionary}.

Recently, some researchers notice the extremely imbalance problem on datasets with imbalance ratio greater than 100 \cite{tang2009svms, triguero2015rosefw}. This context easily leads to a classification task in the Big Data since the positive class is a very small fraction in the dataset, if the dataset is small then we cannot find any meaningful patterns on a few positive data points. To our best knowledge, there are a few studies on this gap of extremely imbalanced big data classification. For example, Sara del Río \cite{del2014use} tested their study on datasets with a largest size is over five million data points or a maximum IR is 74680. This new scenario raises a question about how effectiveness we could achieve with this kind of data. An ideal method need to use any given data points very effective, for both positive and negative data points.

In the traditional imbalance problem, many evaluation measures could not be used in this case independently since they are affected by the majority class. In order to evaluate the classifier, it requires a stronger metric that balance the classes and a commonly used metric for this case is $F_1$ score that can derive from confusion matrix (table \ref{tbl:confusion_matrix}). Based on two other metrics, Precision and Recall, the $F_1$ score is the harmonic average of them and could be computed as formula \ref{eq:f1}. In this study we propose to use the $F_1$ score as a main measure.

\begin{table}[h]
\centering
\caption{Confusion matrix}
\label{tbl:confusion_matrix}
\begin{tabular}{lll}
\hline
               & predicted positives & predicted negatives \\ \hline
real positives & True positive (TP)  & False negative (FN) \\ \hline
real negatives & False positive (FP) & True negative (TN)  \\ \hline
\end{tabular}
\end{table}

\begin{equation}
\label{eq:precision}
Precision = \dfrac{TP}{TP + FP}
\end{equation}

\begin{equation}
\label{eq:recall}
Recall = \dfrac{TP}{TP + FN}
\end{equation}

\begin{equation}
\label{eq:f1}
F_1 = 2 \cdot \dfrac{Precision \cdot Recall}{Precision + Recall}
\end{equation}

\section{Proposed analysis}

Based on the question posed in the previous section, we propose another way to build samples that effectively use all of the data points in the dataset once time. Using the under-sampling method, all negative data points will be randomly divided into several dis-joint same-size subsets then each subset will be merged back with all positive data points then it is used to train a classifier. After that, we combine these classifiers by the voting ensemble as a final prediction. Suppose a dataset $D$ has $N$ data points with imbalance ratio is $IR$, let $K$ be a number of subsets, our proposed approach is summarized by Algorithm \ref{alg:ksus}, in which we define $D^{major}$ and $D^{minor}$ are the majority and minority classes in the dataset $D$ respectively and $n = \begin{bmatrix} \dfrac{ \begin{vmatrix} D^{major} \end{vmatrix} }{ K } \end{bmatrix}$ is a size of a sampled subset. In the main part of Algorithm \ref{alg:ksus},  in this study we mainly use Random Forest because of its robustness and good performance.

\begin{algorithm}[h]
\caption{$K$-Segments Under Bagging ($K$-SUB)}
\label{alg:ksus}

Inputs:
\newline
$D^{major}$ is the majority class in the dataset $D$
\newline
$D^{minor}$ is the minority class in the dataset $D$
\newline
$K$ is the number of subsets

Procedure:
\begin{algorithmic}[1]
\For {$k = 1$ to $K$}:
  \State Draw a subset $K^{major}_k$ without replacement from the majority class with the size $n = \begin{bmatrix} \dfrac{ \begin{vmatrix} D^{major} \end{vmatrix} }{ K } \end{bmatrix}$
  \State Let $K_k$ is a merged subset of $K^{major}_k$ with all data points in minority class $D^{minor}$
  \State Train a classifier $f^k$ by applying Random Forest algorithm to the subset $K_k$
\EndFor
\State Combine all $f_k$ into an aggregated model $F$
\State Return F
\end{algorithmic}


\end{algorithm}


To illustrate the effectiveness of our proposed algorithm, we apply it on many datasets with various aspects, e.g. from small to large datasets, from low to very high imbalance ratio and moreover we try to do our test with several values of $K$, from 3, 5, 10 to 20. We run a stratified 5-fold cross-validation then compute the metric $F_1$ score as an average of five folds and the running time is also reported in seconds. This test is carried out in one single machine with 24 cores and 128 gigabytes in memory. All details of the final results are summarized in the table \ref{tbl:summary_results}.


\begin{table*}[]
\caption{Summary of datasets and the experimental results}
\label{tbl:summary_results}
\resizebox{\textwidth}{!}{%
\begin{tabular}{|c|c|c|c|c|c|c|c|c|c|c|c|c|c|}
\hline
\multirow{2}{*}{dataset} & \multirow{2}{*}{IR} & \multirow{2}{*}{sample} & \multirow{2}{*}{feature} & \multicolumn{2}{c|}{UB}  & \multicolumn{2}{c|}{3-SUB} & \multicolumn{2}{c|}{5-SUB} & \multicolumn{2}{c|}{10-SUB} & \multicolumn{2}{c|}{20-SUB} \\ \cline{5-14} 
                         &                     &                         &                          & F1              & time   & F1                & time   & F1                & time   & F1                 & time   & F1            & time        \\ \hline
ecoli                    & 8                   & 336                     & 7                        & \textbf{0.5758} & 1      & 0.5669            & 1      & 0.4850            & 2      & 0.4563             & 5      & 0.3953        & 9           \\ \hline
spectrometer             & 10                  & 531                     & 93                       & 0.7642          & 1      & \textbf{0.8411}   & 1      & 0.8248            & 2      & 0.7933             & 4      & 0.6601        & 10          \\ \hline
car\_eval\_34            & 11                  & 1728                    & 21                       & \textbf{0.7173} & 2      & 0.4445            & 1      & 0.2418            & 2      & 0.2380             & 5      & 0.2081        & 9           \\ \hline
isolet                   & 11                  & 7797                    & 617                      & 0.6084          & 5      & \textbf{0.7702}   & 3      & 0.7326            & 4      & 0.5984             & 7      & 0.4710        & 13          \\ \hline
libras\_move             & 14                  & 360                     & 90                       & 0.6797          & 2      & \textbf{0.7389}   & 1      & 0.6473            & 2      & 0.4036             & 4      & 0.1717        & 9           \\ \hline
thyroid\_sick            & 15                  & 3772                    & 52                       & 0.7551          & 2      & 0.8023            & 2      & \textbf{0.8261}   & 3      & 0.8112             & 5      & 0.6708        & 10          \\ \hline
oil                      & 21                  & 937                     & 49                       & \textbf{0.3083} & 3      & 0.0941            & 1      & 0.0829            & 2      & 0.0848             & 4      & 0.0841        & 9           \\ \hline
car\_eval\_4             & 25                  & 1728                    & 21                       & \textbf{0.6078} & 4      & 0.4413            & 1      & 0.2183            & 2      & 0.1623             & 5      & 0.1481        & 10          \\ \hline
letter\_img              & 26                  & 20000                   & 16                       & 0.6133          & 5      & \textbf{0.8681}   & 3      & 0.8381            & 4      & 0.8059             & 9      & 0.7018        & 15          \\ \hline
webpage                  & 34                  & 344780                  & 300                      & 0.3783          & 14     & 0.1842            & 5      & 0.3130            & 6      & \textbf{0.4496}    & 12     & 0.1616        & 18          \\ \hline
mammography              & 42                  & 11183                   & 6                        & 0.4033          & 5      & 0.6478            & 1      & \textbf{0.6606}   & 1      & 0.6316             & 3      & 0.5221        & 5           \\ \hline
protein\_homo            & 111                 & 145751                  & 74                       & 0.4877          & 24     & 0.7908            & 7      & 0.7953            & 7      & \textbf{0.8125}    & 12     & 0.8113        & 15          \\ \hline
10\% kddcup R2L          & 443                 & 494021                  & 41                       & 0.2396          & 643    & 0.8945            & 44     & \textbf{0.9350}   & 46     & 0.8803             & 90     & 0.7176        & 101         \\ \hline
fraud\_detection         & 577                 & 284807                  & 29                       & 0.2341          & 148    & \textbf{0.8113}   & 30     & 0.6231            & 30     & 0.5567             & 55     & 0.3304        & 62          \\ \hline
kdd SF PROBE             & 7988                & 703067                  & 30                       & 0.0126          & 3980   & 0.7895            & 80     & 0.8034            & 79     & \textbf{0.8242}    & 156    & 0.7176        & 156         \\ \hline
10\% kdd U2R             & 9499                & 494021                  & 41                       & 0.0075          & 7008   & \textbf{0.6034}   & 38     & \textbf{0.6034}   & 39     & 0.5782             & 79     & 0.5291        & 90          \\ \hline
kdd U2R                  & 94199               & 4940219                 & 41                       & 0.0007          & 112285 & 0.3741            & 359    & 0.4571            & 343    & \textbf{0.5617}    & 737    & 0.4916        & 780         \\ \hline
\end{tabular}%
}
\end{table*}

The results in bold indicate the best $F_1$ scores from the Under Bagging (UB) algorithm and our method $K$-Segments Under Bagging ($K$-SUB) with multiple $K$ values. With datasets have low imbalance ratio (e.g. IR $<$ 50) the Under Bagging algorithm could handle the problem, and our method has also comparative results. However, from a high to extreme imbalance cases, i.e. IR $>=$ 50, our method outperform the Under Bagging in all cases with any value of our chosen $K$ values. Furthermore, the running time is dramatically slow for the case of Under Bagging algorithm, but with our method it only takes us few minutes in the same machine. Specially in the two biggest datasets with highest $IR$, the $F_1$ scores of the Under Bagging tend to become zero but our method still keeps the high $F_1$ scores. In the range of $K$ values selected above, we observe the case $K = 20$ is not the best one, so it suggests that only a small enough value of $K$ may be a good choice in our proposed analysis.

\section{Conclusion}

In this study, we have presented the $K$-Segments Under Bagging algorithm in the attempt to tackle the problem of extremely imbalanced data classification. The experimental results show that in the case of extremely imbalanced data, our method not only outperforms the previous method but also runs very fast. While in case of low imbalanced data, with the suitable $K$ value, the $K$-SUB algorithm is still useful. Furthermore, it can be observed experimentally that we may obtain the good results for the small enough values of $K$, which suggests that in real applications we only have to tune the parameter $K$ in a small range of value. With this new and challenging scenario, we have shown that the simplified combination of undersampling technique and ensemble learning is able to give better results instead of using other complicated methods addressed in the past. In the future work, we want to investigate deeper on this approach as well as other related issues.

\chapter{Solve fraud detection problem by using graph based learning methods}

\ifpdf
    \graphicspath{{Chapter1/Figs/Raster/}{Chapter1/Figs/PDF/}{Chapter1/Figs/}}
\else
    \graphicspath{{Chapter1/Figs/Vector/}{Chapter1/Figs/}}
\fi

The credit cards’ fraud transactions detection is the important problem in machine learning field. To detect the credit cards’ fraud transactions help reduce the significant loss of the credit cards’ holders and the banks. To detect the credit cards’ fraud transactions, data scientists normally employ the un-supervised learning techniques and supervised learning technique. In this paper, we employ the graph p-Laplacian based semi-supervised learning methods combined with the under-sampling technique such as Cluster Centroids to solve the credit cards’ fraud transactions detection problem. Experimental results show that that the graph p-Laplacian semi-supervised learning methods outperform the current state of art graph Laplacian based semi-supervised learning method ($p$ = 2). In the scope of this thesis, some necessary knowledge will not be presented and the reader may find them in \citep{tran2012application, latouche2015graphs}.

\section{Introduction}

While purchasing online, the transactions can be done by using credit cards that are issued by the bank. In this case, if the cards or cards’ details are stolen, the fraud transactions can be easily carried out. This will lead to the significant loss of the cardholder or the bank. In order to detect credit cards’ fraud transactions, data scientists employ a lot of machine learning techniques. To the best of our knowledge, there are two classes of machine learning techniques used to detect credit cards’ fraud transactions which are un-supervised learning techniques and supervised learning techniques. The un-supervised learning techniques used to detect credit cards’ fraud transactions are k-means clustering technique \citep{goldstein2016comparative}, k-nearest neighbors technique \citep{goldstein2016comparative}, Local Outlier Factor technique \citep{goldstein2016comparative}, to name a few. The supervised learning techniques used to detect credit cards’ fraud transactions are Hidden Markov Model technique \citep{singh2012survey}, neural network technique \citep{nune2015novel}, Support Vector Machine technique \citep{csahin2011detecting}, to name a few.

To the best of our knowledge, the graph based semi-supervised learning techniques \citep{shin2007graph} have not been applied to the credit cards’ fraud transactions detection problem. In this paper, we will apply the un-normalized graph p-Laplacian based semi-supervised learning technique \citep{tran2015normalized, tran2017normalized} combined with the under-sampling technique to the credit cards’ fraud transactions detection problem.

We will organize the paper as follows: Section 2 will introduce the preliminary notations and definitions used in this paper. Section 3 will introduce the definitions of the gradient and divergence operators of graphs. Section 4 will introduce the definition of Laplace operator of graphs and its properties. Section 5 will introduce the definition of the curvature operator of graphs and its properties. Section 6 will introduce the definition of the p-Laplace operator of graphs and its properties. Section 7 will show how to derive the algorithm of the un-normalized graph p-Laplacian based semi-supervised learning method from regularization framework. In section 8, we will compare the accuracy performance measures of the un-normalized graph Laplacian based semi-supervised learning algorithm (i.e. the current state of art graph based semi-supervised learning method) combined with the under-sampling technique such as Cluster Centroids technique \citep{yen2009cluster} and the un-normalized graph p-Laplacian based semi-supervised learning algorithms combined with Cluster Centroids technique \citep{yen2009cluster}. Section 9 will conclude this paper and the future direction of researches.

\section{Preliminary notations and definitions}

Given a graph  $G = (V, E, W)$  where  $V$  is a set of vertices with  $\vert V \vert = n$, $E \subseteq V \ast V$ is a set of edges and  $W$ is a  $n\ast n$  similarity matrix with elements  $w_{ij} \geq 0  \left( 1 \leq i,j \leq n \right)  $.

Also, please note that  $ w_{ij}=w_{ji} $ .

The degree function  \( d:V \rightarrow R^{+} \)  is

\begin{equation}
d_{i}= \sum _{j \sim i}^{}w_{ij}
\end{equation}

where  \( j \sim i \)  is the set of vertices adjacent with \textit{i}.

Define  \( D=diag \left( d_{1},d_{2}, \ldots ,d_{n} \right)  \) .

The inner product on the function space  \( R^{V} \)  is

\begin{equation}
<f,g>_{V}= \sum _{i \in V}^{}f_{i}g_{i}
\end{equation}

Also, define an inner product on the space of functions  \( R^{E} \)  on the edges

\begin{equation}
<F,G>_{E} = \sum _{ \left( i,j \right)  \in E}^{}F_{ij}G_{ij}
\end{equation}

Here let  \( H \left( V \right) = \left( R^{V},<.,.>_{V} \right)  \)  and  \( H \left( E \right) = \left( R^{E},<.,.>_{E} \right)  \)  be the Hilbert space real-valued functions defined on the vertices of the graph \textit{G} and the Hilbert space of real-valued functions defined in the edges of \textit{G} respectively.

\section{Gradient and Divergence Operators}

We define the gradient operator  \( d:H \left( V \right)  \rightarrow H \left( E \right)  \)  to be\par

\begin{equation}
\left( df \right) _{ij}=\sqrt[]{w_{ij}} \left( f_{j}-f_{i} \right)
\end{equation}

where  \( f:V \rightarrow R \)  be a function of  \( H \left( V \right)  \) .

We define the divergence operator  \( div:H \left( E \right)  \rightarrow H \left( V \right)  \)  to be

\begin{equation}
<df,F>_{H \left( E \right) }=<f,-divF>_{H \left( V \right) },
\end{equation}

where  \( f \in H \left( V \right) ,F \in H \left( E \right)  \) .

Thus, we have

\begin{equation}
\left( divF \right) _{j}= \sum _{i \sim j}^{}\sqrt[]{w_{ij}} \left( F_{ji}-F_{ij} \right)
\end{equation}

\section{Laplace operator}

We define the Laplace operator  \(  \Delta :H \left( V \right)  \rightarrow H \left( V \right)  \)  to be

\begin{equation}
\Delta f=-\frac{1}{2}div \left( df \right)
\end{equation}

Thus, we have

\begin{equation}
\left(  \Delta f \right) _{j}=d_{j}f_{j}- \sum _{i \sim j}^{}w_{ij}f_{i}
\end{equation}

The graph Laplacian is a linear operator. Furthermore, the graph Laplacian is self-adjoint and positive semi-definite. 

Let  \( S_{2} \left( f \right) =< \Delta f,f> \) , we have the following \textbf{theorem 1}

\begin{equation}
D_{f}S_{2}=2 \Delta f
\end{equation}

The proof of the above theorem can be found from \citep{tran2015normalized, tran2017normalized}.

\section{Curvature operator}

We define the curvature operator  \(  \kappa :H \left( V \right)  \rightarrow H \left( V \right)  \)  to be

\begin{equation}
\kappa f=-\frac{1}{2}div \left( \frac{df}{ \vert  \vert df \vert  \vert } \right)
\end{equation}

Thus, we have

\begin{equation}
\label{eq:kf_j}
\left(  \kappa f \right) _{j}=\frac{1}{2} \sum _{i \sim j}^{}w_{ij} \left( \frac{1}{ \Vert d_{i}f \Vert }+\frac{1}{ \Vert d_{j}f \Vert } \right)  \left( f_{j}-f_{i} \right)
\end{equation}

From the above formula, we have

\begin{equation}
d_{i}f= \left(  \left( df \right) _{ij}:j \sim i \right) ^{T}
\end{equation}

The local variation of \textit{f} at \textit{i} is defined to be

\begin{equation}
\Vert d_{i}f \Vert =\sqrt[]{ \sum _{j \sim i}^{} \left( df \right) _{ij}^{2}}=\sqrt[]{ \sum _{j \sim i}^{}w_{ij} \left( f_{j}-f_{i} \right) ^{2}}
\end{equation}

To avoid the zero denominators in \ref{eq:kf_j}, the local variation of \textit{f} at \textit{i} is defined to be

\begin{equation}
\Vert d_{i}f \Vert =\sqrt[]{ \sum _{j \sim i}^{} \left( df \right) _{ij}^{2}+ \epsilon }
\end{equation}

where  \(  \epsilon =10^{-10} \) .

The graph curvature is a non-linear operator.    

Let  \( S_{1} \left( f \right) = \sum _{i}^{} \Vert d_{i}f \Vert  \) , we have the following \textbf{theorem 2}

\begin{equation}
D_{f}S_{1}= \kappa f
\end{equation}

The proof of the above theorem can be found from \citep{tran2015normalized, tran2017normalized}.

\section{p-Laplace operator}

We define the p-Laplace operator  \(  \Delta _{p}:H \left( V \right)  \rightarrow H \left( V \right)  \)  to be

\begin{equation}
\Delta _{p}f=-\frac{1}{2}div \left(  \Vert df \Vert ^{p-2}df \right)
\end{equation}

Thus, we have

\begin{equation}
\left(  \Delta _{p}f \right) _{j}=\frac{1}{2} \sum _{i \sim j}^{}w_{ij} \left(  \Vert d_{i}f \Vert ^{p-2}+ \Vert d_{j}f \Vert ^{p-2} \right)  \left( f_{j}-f_{i} \right)
\end{equation}

Let  \( S_{p} \left( f \right) =\frac{1}{p} \sum _{i}^{} \Vert d_{i}f \Vert ^{p} \) , we have the following \textbf{theorem 3}\ \ \ \  \par

\begin{equation}
D_{f}S_{p}=p \Delta _{p}f
\end{equation}

\section{Discrete regularization on graphs and credit cards’ fraud transactions detection problems}

Given a transaction network \textit{G=(V,E)}. \textit{V} is the set of all transactions in the network and \textit{E} is the set of all possible interactions between these transactions. Let \textit{y} denote the initial function in \textit{H(V)}.  \( y_{i} \) \textit{ }can be defined as follows

$$
y_{i} = \begin{cases}

 & 1 \text{ if transaction } i \text{ is the fraud transaction} \\ 
 & -1 \text{ if transaction } i \text{ is the normal transaction} \\
 & 0 \text{ otherwise} \\

\end{cases}
$$

Our goal is to look for an estimated function \textit{f} in \textit{H(V)} such that \textit{f} is not only smooth on \textit{G} but also close enough to an initial function \textit{y}. Then each transaction \textit{i} is classified as  \( sign \left( f_{i} \right)  \) . This concept can be formulated as the following optimization problem

\begin{equation}
\label{eq:opt}
argmin_{f \in H \left( V \right) } \{ S_{p} \left( f \right) +\frac{ \mu }{2} \Vert f-y \Vert ^{2} \} 
\end{equation}

The first term in \ref{eq:opt} is the smoothness term. The second term is the fitting term. A positive parameter  \(  \mu  \)  captures the trade-off between these two competing terms.

\subsection{p-smoothness}

For any number \textit{p}, the optimization problem \ref{eq:opt} is

\begin{equation}
\label{eq:p_smoothness}
argmin_{f \in H \left( V \right) } \{ \frac{1}{p} \sum _{i}^{} \Vert d_{i}f \Vert ^{p}+\frac{ \mu }{2} \Vert f-y \Vert ^{2} \}
\end{equation}

By theorem 3, we have

\textbf{Theorem 4:} The solution of \ref{eq:p_smoothness} satisfies

\begin{equation}
\label{eq:opt_satisfies}
\Delta _{p}f+ \mu  \left( f-y \right) = 0
\end{equation}

The \textit{p-Laplace} operator is a non-linear operator; hence we do not have the closed form solution of equation \ref{eq:opt_satisfies}. Thus, we have to construct iterative algorithm to obtain the solution. From \ref{eq:opt_satisfies}, we have

\begin{equation}
\label{eq:opt_ext}
\frac{1}{2} \sum _{i \sim j}^{}w_{ij} \left(  \Vert d_{i}f \Vert ^{p-2}+ \Vert d_{j}f \Vert ^{p-2} \right)  \left( f_{j}-f_{i} \right) + \mu  \left( f_{j}-y_{j} \right) = 0
\end{equation}

Define the function  \( m:E \rightarrow R \) by

\begin{equation}
m_{ij}=\frac{1}{2}w_{ij} \left(  \Vert d_{i}f \Vert ^{p-2}+ \Vert d_{j}f \Vert ^{p-2} \right)
\end{equation}

Then equation \ref{eq:opt_ext} which is 

$$\sum _{i \sim j}^{}m_{ij} \left( f_{j}-f_{i} \right) + \mu  \left( f_{j}-y_{j} \right) =0 $$

can be transformed into

\begin{equation}
\left(  \sum _{i \sim j}^{}m_{ij}+ \mu  \right) f_{j}= \sum _{i \sim j}^{}m_{ij}f_{i}+ \mu y_{j}
\end{equation}

Define the function  \( p:E \rightarrow R \)  by

\begin{equation}
p_{ij} = \begin{cases}
  \frac{m_{ij}}{ \sum _{i \sim j}m_{ij}+ \mu } \text{ if } i \neq j \\
  \frac{ \mu }{ \sum _{i \sim j}m_{ij}+ \mu } \text{ if } i=j \\
\end{cases}
\end{equation}

Then

\begin{equation}
f_{j}= \sum _{i \sim j}^{}p_{ij}f_{i}+p_{jj}y_{j}
\end{equation}

Thus we can consider the iteration 

$$f_{j}^{ \left( t+1 \right) }= \sum _{i \sim j}^{}p_{ij}^{ \left( t \right) }f_{i}^{ \left( t \right) }+p_{jj}^{ \left( t \right) }y_{j} \text{ for all } j \in V $$

to obtain the solution of \ref{eq:p_smoothness}.

\section{Experiments and results}

\textbf{Datasets}

In this paper, we use the transaction dataset available from \citep{dal2015calibrating}. This dataset contains 284,807 transactions. Each transaction has 30 features. In the other words, we are given transaction data matrix ( \( R^{284807\ast30} \) ) and the annotation (i.e. the label) matrix ( \( R^{284807\ast1} \) ). The ratio between the number of fraud transactions and the number of normal transactions is 0.00173. Hence we easily recognize that this is the imbalanced classification problem. In order to solve this imbalanced classification problem, we initially apply the under-sampling technique which is the Cluster Centroid technique \citep{yen2009cluster} to this imbalanced dataset. Then we have that the ratio between the number of fraud transactions and the number of normal transactions is 0.4. In the other words, we are given the \textbf{new transaction data} matrix ( \( R^{1722\ast30} \) ) and the annotation (i.e. the label) matrix ( \( R^{1722\ast1} \) ).\par

Then we construct the similarity graph from the transaction data. The similarity graph used in this paper is the k-nearest neighbor graph: Transaction \textit{i} is connected with transaction \textit{j} if transaction \textit{i} is among the k-nearest neighbor of transaction \textit{j} or transaction \textit{j} is among the k-nearest neighbor of transaction \textit{i}.\  \textit{ } \par

In this paper, the similarity function is the Gaussian similarity function

$$
s \big( T(i,:), T(j,:) \big) = exp \bigg( - \frac{ d \big(T(i,:), T(j,:) \big) }{t} \bigg)
$$

In this paper, $t$ is set to 0.1 and the 5-nearest neighbor graph is used to construct the similarity graph from the \textbf{new transaction data}.\ \ \ \ \  \par

\textbf{Experimental Results}

In this section, we experiment with the above proposed un-normalized graph p-Laplacian methods with \textit{p=1, 1.1, 1.2, 1.3, 1.4, 1.5, 1.6, 1.7, 1.8, 1.9} and the current state of the art method (i.e. the un-normalized graph Laplacian based semi-supervised learning method \textit{p=2}) in terms of classification accuracy performance measure. The accuracy performance measure Q is given as follows

$$Q=\frac{True Positive+True Negative}{True Positive+True Negative+False Positive+False Negative}$$

The \textbf{new transaction data} is divided into two subsets: the training set and the testing set. The training set contains 1,208 transactions. The testing set contains 514 transactions. The parameter  \(  \mu  \)  is set to 1.

The accuracy performance measures of the above proposed methods and the current state of the art method is given in the following table 1


\begin{table}[]
\centering
\caption{The comparison of accuracies of proposed methods with different p-values}
\label{tbl:accuracy_performance_measures}
\begin{tabular}{|c|c|}

\hline
\multicolumn{2}{|c|}{Accuracy Performance Measures ($\%$ )} \\
\hline
\multicolumn{1}{|c}{p=1} & 
\multicolumn{1}{|c|}{88.52} \\
\hline
\multicolumn{1}{|c}{p=1.1} & 
\multicolumn{1}{|c|}{88.52} \\
\hline
\multicolumn{1}{|c}{p=1.2} & 
\multicolumn{1}{|c|}{88.52} \\
\hline
\multicolumn{1}{|c}{p=1.3} & 
\multicolumn{1}{|c|}{88.52} \\
\hline
\multicolumn{1}{|c}{p=1.4} & 
\multicolumn{1}{|c|}{88.52} \\
\hline
\multicolumn{1}{|c}{p=1.5} & 
\multicolumn{1}{|c|}{88.52} \\
\hline
\multicolumn{1}{|c}{p=1.6} & 
\multicolumn{1}{|c|}{88.52} \\
\hline
\multicolumn{1}{|c}{p=1.7} & 
\multicolumn{1}{|c|}{88.52} \\
\hline
\multicolumn{1}{|c}{p=1.8} & 
\multicolumn{1}{|c|}{88.52} \\
\hline
\multicolumn{1}{|c}{p=1.9} & 
\multicolumn{1}{|c|}{88.52} \\
\hline
\multicolumn{1}{|c}{p=2} & 
\multicolumn{1}{|c|}{88.33} \\
\hline

\end{tabular}
\end{table}


From the above table, we easily recognized that the un-normalized graph p-Laplacian semi-supervised learning methods outperform the current state of art method. The results from the above table show that the un-normalized graph p-Laplacian semi-supervised learning methods are at least as good as the current state of the art method (\textit{p=2}) but often lead to better classification accuracy performance measures.

\section{Conclusions}

We have developed the detailed regularization frameworks for the un-normalized graph p-Laplacian semi-supervised learning methods applying to the credit cards’ fraud transactions detection problem. Experiments show that the un-normalized graph p-Laplacian semi-supervised learning methods are at least as good as the current state of the art method (i.e. \textit{p=2}) but often lead to significant better classification accuracy performance measures. 

In the future, we will develop the detailed regularization frameworks for the un-normalized hypergraph p-Laplacian semi-supervised learning methods and will apply these methods to this credit cards’ fraud transactions detection problem.

\chapter{Conclusion and Future works}

\ifpdf
    \graphicspath{{Chapter1/Figs/Raster/}{Chapter1/Figs/PDF/}{Chapter1/Figs/}}
\else
    \graphicspath{{Chapter1/Figs/Vector/}{Chapter1/Figs/}}
\fi

Fraud detection is a challenging and complex problem in many real-world applications, particularly credit card fraud detection problem. This thesis investigates on how to use Machine Learning and Data Science to address some of the issues in this problem. This chapter summarizes the main results of this thesis, discusses open issues and presents our future work.

Chapter 3 is the survey on various challenges of the fraud detection problem, with each challenge we also presented some most suitable solutions. After they are analyzed, we have proposed the most simple and effective strategy to build the Fraud Detection Pipeline that could be used as the backbone of one fraud detection system. Based on our Fraud Detection Pipeline, we also suggest many ways to extend or upgrade it. With this comprehensive survey, we want to write more details and clearly to contribute to the community the most recent works in the fraud detection problem.

Using our Fraud Detection Pipeline, which requires to update all models in the system every time frame (e.g daily), in chapter 4 we presented the mechanism to predict the improvement if we update a model, then with the pre-defined threshold we could decide which model need to update. The result shows that our system reduces the number of update times significantly, i.e 80\%. In future work, we want to replace the pre-defined threshold with a dynamic threshold that can make our system more stable and also not-sensitive with our parameter.

In the Fraud Detection Pipeline, we have proposed to use undersampling technique and ensemble learning for the credit card fraud detection dataset, which is not only huge but also very high imbalanced. In chapter 5, we presented the study of this combination on the case of extremely imbalanced big data classification and the result shows that our approach is very effective and promising if the dataset is bigger or more imbalance. In future work, we want to investigate deeper on this new gap of two difficult problems, e.g feature selection on the extremely imbalanced big data classification.

Most of the Machine Learning algorithms are applied to the fraud detection problem, but the graph-based learning has not been considered. In chapter 6, we used the graph p-Laplacian based semi-supervised learning with undersampling on this problem and the result shows that it outperforms the current state of the art graph Laplacian based semi-supervised method. Finding the relationship of the fraud network is one of the hardest tasks in the fraud detection problem and the graph-based learning attracts more attention recently. In future work, we want to apply the graph-based learning to detect the fraud networks in our data.


\begin{spacing}{0.9}


\bibliographystyle{apalike}
\cleardoublepage
\bibliography{thesis} 



\end{spacing}


\begin{appendices} 

\end{appendices}

\printthesisindex 

\end{document}